\documentclass[10pt,twocolumn,letterpaper]{article}

\usepackage[pagenumbers]{cvpr} 

\usepackage{graphicx}

\usepackage[dvipsnames]{xcolor}

\usepackage{relsize}

\usepackage{multirow}
\usepackage{xcolor, colortbl}

\definecolor{darkred}{rgb}{0.7,0.1,0.1}
\definecolor{darkgreen}{rgb}{0.1,0.6,0.1}
\definecolor{cyan}{rgb}{0.7,0.0,0.7}
\definecolor{otherblue}{rgb}{0.1,0.4,0.8}
\definecolor{maroon}{rgb}{0.76,.13,.28}
\definecolor{burntorange}{rgb}{0.81,.33,0}
\definecolor{apricot}{rgb}{0.98, 0.81, 0.69}

\usepackage{pifont}

\usepackage{caption}
\usepackage{booktabs}
\usepackage{arydshln}

\newcommand{\flowchef}{\textbf{\texttt{FlowChef}}}

\newcommand{\nresults}[1]{\textsubscript{{\color{red}\textbf{#1}}}}

\newcolumntype{C}[1]{>{\centering\arraybackslash}p{#1}}

\usepackage{amsthm}

\newtheorem{theorem}{Theorem}[section] 
\newtheorem{proposition}[theorem]{Proposition} 
\newtheorem{lemma}[theorem]{Lemma} 

\theoremstyle{definition} 

\usepackage[linesnumbered,ruled,vlined]{algorithm2e}
\usepackage{booktabs}
\usepackage{multirow}
\usepackage{graphicx}
\usepackage{caption}
\usepackage{amsmath}
\usepackage{tikz}
\usetikzlibrary{decorations.pathreplacing} 

\usepackage[normalem]{ulem}

\usepackage{multibib}
\newcites{App}{References for the Appendix}

\definecolor{cvprblue}{rgb}{0.21,0.49,0.74}
\usepackage[pagebackref,breaklinks,colorlinks,allcolors=cvprblue]{hyperref}

\def\paperID{****} 
\def\confName{CVPR}
\def\confYear{2025}

\title{Steering Rectified Flow Models in the Vector Field \\for Controlled Image Generation}

\author{Maitreya Patel\textsuperscript{$\spadesuit$}, \quad Song Wen\textsuperscript{$\diamondsuit$}, \quad Dimitris N. Metaxas\textsuperscript{$\diamondsuit$}, \quad Yezhou Yang\textsuperscript{$\spadesuit$}\\
\textsuperscript{$\spadesuit$}
Arizona State University \quad \textsuperscript{$\diamondsuit$}Rutgers University\\
{\tt\small \{maitreya.patel, yz.yang\}@asu.edu \quad \quad  \{song.wen, dnm\}@rutgers.edu }
}

\begin{document}

\twocolumn[{%
\renewcommand\twocolumn[1][]{#1}%
\maketitle

\vspace{-0.5cm}

\centering
\captionsetup{type=figure}
\includegraphics[width=\linewidth]{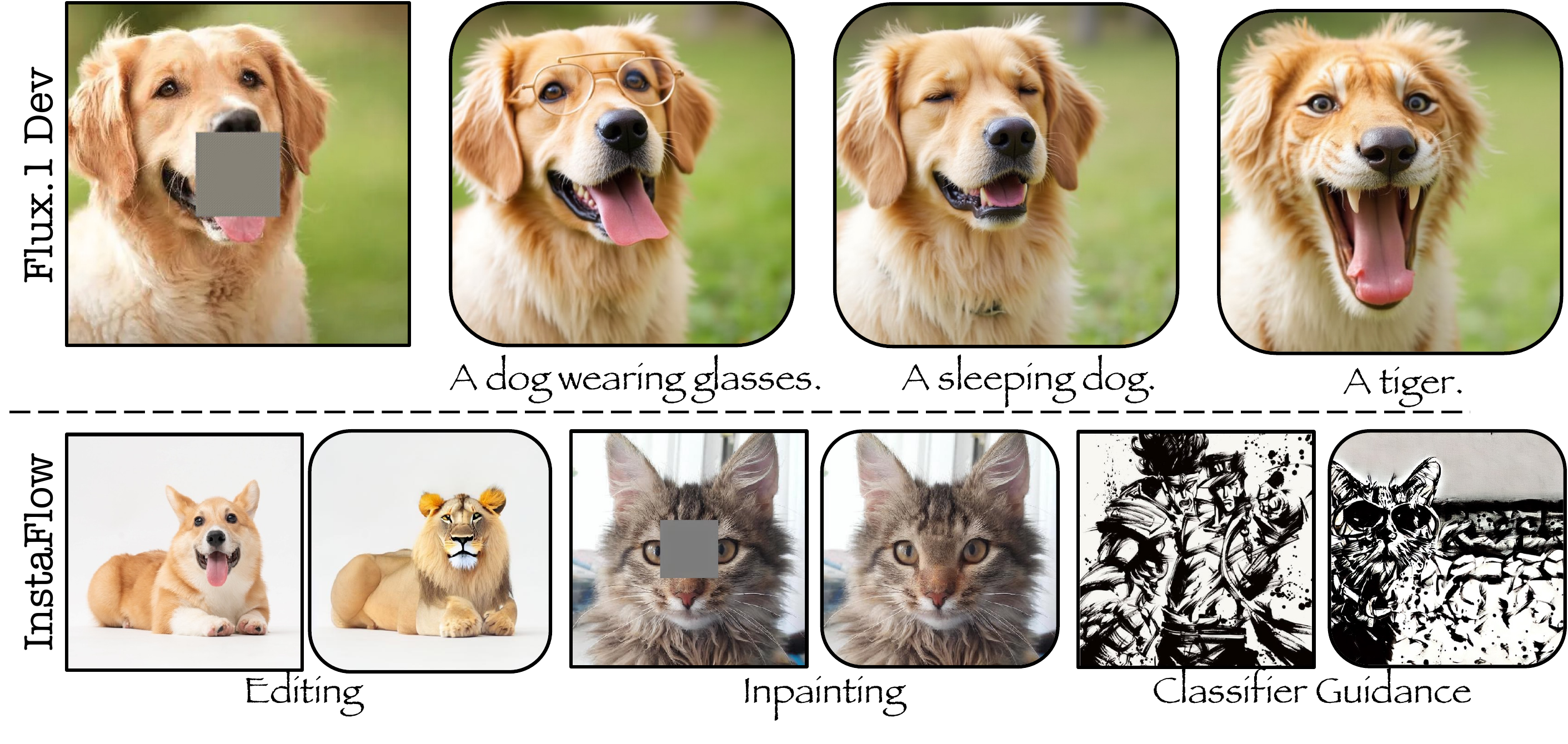}
\vspace{-1.0cm}
\captionof{figure}{\flowchef~steers the trajectory of Rectified Flow Models during inference to tackle linear inverse problems, image editing, and classifier guidance. We extend \flowchef~to SOTA models like Flux and InstaFlow, enabling gradient- and inversion-free control for efficient, controlled image generation.}
\label{fig:teaser}
\vspace{3pt}
}]

\begin{abstract}

Diffusion models (DMs) excel in photorealism, image editing, and solving inverse problems, aided by classifier-free guidance and image inversion techniques. 
However, rectified flow models (RFMs) remain underexplored for these tasks. 
Existing DM-based methods often require additional training, lack generalization to pretrained latent models, underperform, and demand significant computational resources due to extensive backpropagation through ODE solvers and inversion processes. 
In this work, we first develop a theoretical and empirical understanding of the vector field dynamics of RFMs in efficiently guiding the denoising trajectory. 
Our findings reveal that we can navigate the vector field in a deterministic and gradient-free manner. 
Utilizing this property, we propose \flowchef, which leverages the vector field to steer the denoising trajectory for controlled image generation tasks, facilitated by gradient skipping.  
\flowchef\ is a unified framework for controlled image generation that, for the first time, simultaneously addresses classifier guidance, linear inverse problems, and image editing without the need for extra training, inversion, or intensive backpropagation.
Finally, we perform extensive evaluations and show that \flowchef\ significantly outperforms baselines in terms of performance, memory, and time requirements, achieving new state-of-the-art results.
Project Page: \url{https://flowchef.github.io}.

\end{abstract}    
\section{Introduction}
\label{sec:introduction}

\begin{figure*}[t]
    \centering
    \includegraphics[width=\textwidth]{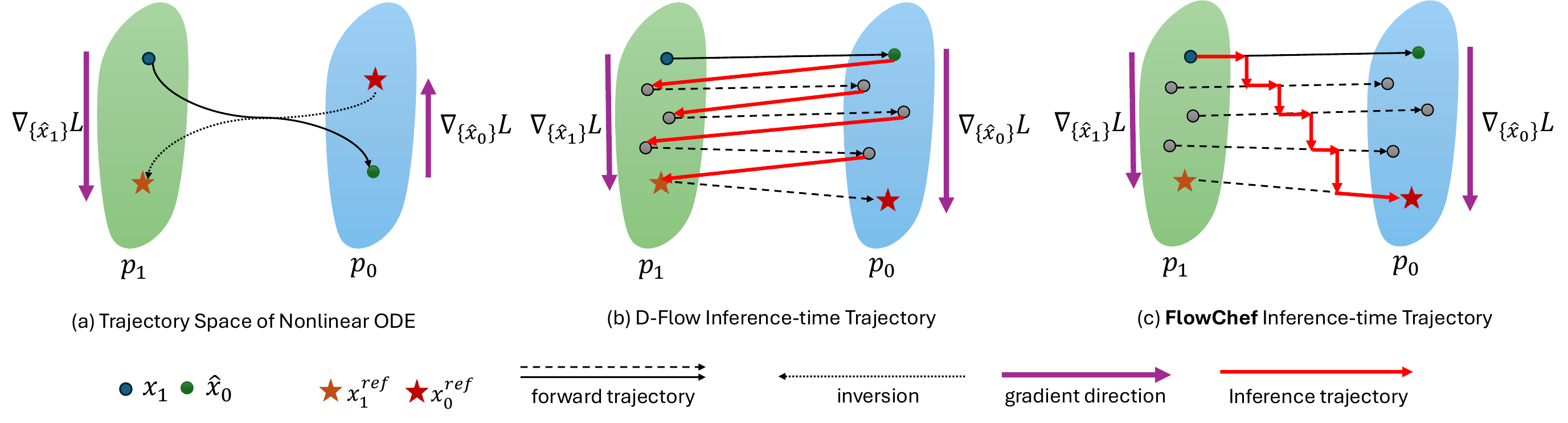}
    \caption{
    \textbf{Motivation behind \flowchef~based on rectified flow models' trajectory space.} Let \( p_1 \sim N(0,I) \) and \( p_0 \) be distributions, with \( x_1 \sim p_1 \) as initial noise, \( x_0^{ref} \) as the target sample, \( \hat{x}_0 \) as the denoised sample from \( x_1 \), and \( x_1^{ref} \) as the specific noise leading to \( x_0^{ref} \). (a) Stochasticity and nonlinear trajectories with crossovers can complicate gradient estimation at each denoising step \( t \). (b) D-Flow (baseline) inference-time trajectory requires the backpropagation through entire denoising steps. (c) Our method \flowchef~enables efficient trajectory steering to guide \( x_t \) along the trajectory towards \( x_0^{ref} \).
}
    \label{fig:method_teaser}
\end{figure*}

Recent advances in diffusion models have led to rapid progress in AI generated content (AIGC), particularly in text-to-image (T2I) and text-to-video (T2V) models across various domains such as entertainment, arts, and design~\cite{rombach2022high, saharia2022photorealistic, esser2024scaling, polyak2024movie, patel2024lambda, singer2023makeavideo}. 
These developments have resulted in remarkable performance in image editing, solving inverse problems, and personalization. 
This progress could be attributed to key advances like latent diffusion models (LDMs)~\cite{rombach2022high} and classifier-free guidance (CFG)~\cite{ho2022classifier}, among other essential components. 
Despite their applicability to various downstream tasks, these models demand increasing computational resources. 
For instance, CFG requires additional unconditional training of the model, while traditional classifier guidance necessitates training noise-aware classifiers~\cite{dhariwal2021diffusion}. 
Similarly, existing approaches for solving inverse problems often require minutes of computation and additional memory overhead~\cite{song2023pseudoinverse, chung2022diffusion, rout2024solving, songsolving, ben-hamu2024dflow}.
Moreover, image editing methods typically involve either inversion or explicit training~\cite{ju2023direct, brack2024ledits++, brooks2023instructpix2pix}.
These limitations can be attributed to the inherent stochasticity of diffusion models, often requiring a higher number of function evaluations (NFEs).

\begin{figure*}[t]
    \centering
    \includegraphics[width=\linewidth]{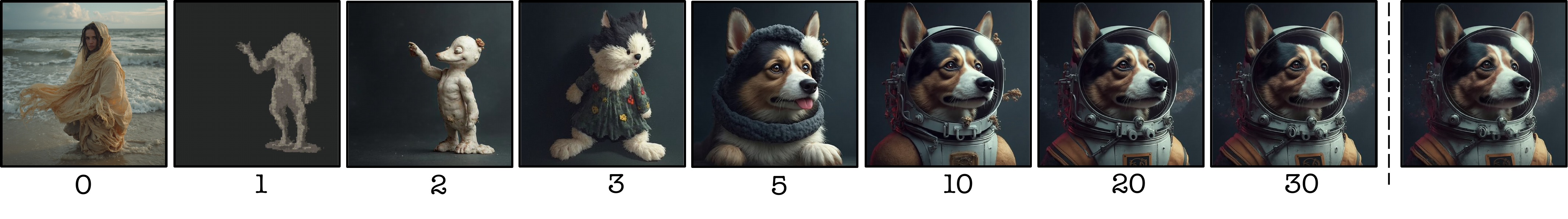}
    \caption{Illustration of impact of guided control step on Flux.1[Dev] with mean squared error as cost function ($\mathcal{L} = || \hat{x}_0 - x_0^{ref} ||^2_2 $). This shows that \flowchef~could guide the rectified flow models on the fly without requiring either the gradients through the Flux model or inversion. Importantly, the convergence speed is slowed down for illustration purposes. }
    \label{fig:algo_hyperparameter}
\end{figure*}

However, the recent introduction of flow-based methods~\cite{lipman2023flow}, especially rectified flow models (RFMs)~\cite{liu2023flow, lee2024improving}, addresses these limitations to some extent by requiring fewer NFEs, depending on the model considered.
Recent works have attempted to solve inverse problems by leveraging this property, focusing mainly on pixel models~\cite{ben-hamu2024dflow, martin2024pnp}. 
While these approaches have improved computational time requirements, they are still not sufficiently efficient, as they require inversion and incur significant memory overhead. 
As a result, they cannot be extended to large state-of-the-art models like Flux or SD3~\cite{esser2024scaling}.

In this paper, we introduce \flowchef, a novel method that significantly enhances controlled image generation by leveraging the unique characteristics of rectified flow models. 
We first standardize the objective of controlled synthesis, unifying various downstream tasks within a single framework. 
By revisiting the ordinary differential equations (ODEs) that govern these models, we analyze their error dynamics both theoretically and empirically. 
We discover that in nonlinear ODEs with stochasticity or trajectory crossovers, error terms emerge that hinder convergence due to inaccuracies in estimating denoised samples or improper gradient approximations (see Figure~\ref{fig:method_teaser}(a)).

Contrary to diffusion models, rectified flow models exhibit straight trajectories and avoid significant trajectory crossovers due to their linear interpolation between noise and data distributions (see Figure~\ref{fig:method_teaser}(b-c)).
We theoretically demonstrate and empirically validate that RFMs can achieve higher convergence rates without additional computational overhead by capitalizing on this key property.
Building on this understanding, we present \flowchef, that proposes to steer the trajectories towards the target in the vector field by gradient skipping (see Figure~\ref{fig:method_teaser}(c)).
This allows us to navigate the vector field in a deterministic manner, akin to a north star guiding sailors across a dark ocean.

We conduct extensive evaluations of \flowchef\ across tasks such as pixel-level classifier guidance, image editing, and classifier-guided style transfer. Our results demonstrate that \flowchef\ not only surpasses baseline methods but does so with greater computational efficiency and without the need for inversion. As illustrated in Figure~\ref{fig:teaser}, \flowchef\ efficiently addresses a variety of tasks. 
For perspective, \flowchef~handles the linear inverse problems within 18 seconds on the latent-space model, while SOTA takes 1-3 minutes per image.
Furthermore, we explore its practical applicability to large-scale models (i.e., Flux) to tackle both linear inverse problems and image editing together without inversion and within 30 NFEs at billions of parameter scales.
Our key contributions can be summarized as follows:
\begin{itemize}
    \item We develop a unified perspective to study rectified flow models theoretically and empirically for a guided, controlled generation.
    \item We introduce \flowchef, the most efficient method to date for guided, controlled generation using RFMs, achieving state-of-the-art performance without requiring inversion or gradient backpropagation through the ODESolver.
    \item We demonstrate \flowchef's superior performance across multiple tasks, including linear inverse problems in both pixel and latent spaces, image editing evaluated on the PIE benchmark~\cite{ju2023direct}, classifier guidance, and through large-scale human preference studies.
\end{itemize}
\section{Related Works}
\label{sec:related_works}

We provide detailed related works, specially diffusion-based methods and conditional sampling, in the Appendix.

\paragraph{Inverse Problems.} This task addresses training-free approaches for solving inverse problems such as in-painting, super resolution, Gaussian de-blurring etc~\cite{daras2024survey}. 
Since Dhariwal et. al. demonstrated that guiding models with classifiers improves image generation quality~\cite{dhariwal2021diffusion}, much of the current literature focuses on diffusion models, particularly pixel-space models~\cite{daras2024survey, song2023pseudoinverse, chung2022diffusion, wu2024principled}. 
However, these models face challenges when scaled to latent-space models, as they are incompatible with off-the-shelf pretrained models and require backpropagation through ODESolvers, which can take at least three minutes per image for satisfactory results~\cite{daras2024survey, rout2024solving, songsolving, rout2024beyond}.
Methods such as MPGD~\cite{hemanifold} attempt to mitigate these issues via manifold correction, but limitations persist, especially with large-scale models. 
Recent work has extended these approaches to ODEs (e.g., OT-ODE) and flow models~\cite{pokle2024trainingfree}. 
D-Flow~\cite{ben-hamu2024dflow}, for instance, optimizes initial noise by differentiating through the full trajectory chain; however, this comes with significant resource demands and is not adaptable to state-of-the-art (SOTA) models like Flux or SD3~\cite{esser2024scaling}. 
In this work, we propose \flowchef, which addresses linear inverse problems in a gradient- and inversion-free manner.

\paragraph{Image Editing.} 
Diffusion-based approaches dominate image editing~\cite{huang2024diffusion}, but they rely heavily on accurate inversion~\cite{ju2023direct, brack2024ledits++, mokady2023null, huberman2024edit}. 
Although inversion-free diffusion methods are faster, they often lack in edit quality~\cite{xu2023inversion, couairon2023diffedit, meng2022sdedit, wu2024turboedit}. 
Despite RFMs being SOTA in text-to-image (T2I) generation, they still lack robust editing capabilities. 
iRDS~\cite{yang2024text} presents an inversion strategy for RFMs, especially InstaFlow~\cite{liu2023instaflow}, but it lacks quality and control. 
Similarly, RectifID~\cite{sun2024rectifid} offers an optimization-based approach to modify the whole trajectory for personalized T2I generation but performs poorly with InstaFlow like straight models.
To the best of our knowledge, we present the first comprehensive solution that enhances RFMs for image editing and extends beyond it that too without significant computational or time overhead.

\paragraph{Concurrent Works.} We note two concurrent works: RF-Inversion~\cite{rout2024semantic} and PnP-Flow~\cite{martin2024pnp}. 
RF-Inversion offers an optimal control-based approach for image inversion and editing, whereas our method generalizes across all controlled generation tasks. 
We demonstrate that inversion is unnecessary, even for RF-Inversion, making RF-Inversion a special case of \flowchef, where starting noise originates from an inverted target image rather than random noise, as in \flowchef. 
PnP-Flow is an inversion- and gradient-free method for inverse problems, but it leads to over-smoothed results and lacks extensibility to image editing.

\section{Preliminaries}
\label{sec:prelim}

Classifier guidance, inversion problems, and image editing involve guiding a model toward a specific target sample or distribution in both pixel and latent spaces. 
However, these tasks are often treated separately in literature. 
Here, we present a unified problem formulation to encompass these downstream tasks, with a focus on rectified flow models.

\subsection{Problem Formulation}

Let $u_\theta: \mathcal{R}^d \times [0, T] \rightarrow \mathcal{R}^d$ represent a pretrained flow model estimating the drift $v = x_1 - x_0$ from $x_t$. 
The denoised sample $\hat{x}_0$ is obtained by integrating the drift $u_\theta$ over time from $t = T$ to $t = 0$, starting from $x_T \sim p_1$. 
With a target sample $x_0^{ref}$, we define a cost function $\mathcal{L}: \mathcal{R}^d \times \mathcal{R}^d \rightarrow \mathcal{R}_+$ that quantifies the cost of aligning $\hat{x}_0$ with $x_0^{ref}$, yielding the optimization problem:

\begin{equation} 
    \min_{\{\hat{x}_t\}_{t=0}^T} \quad \mathcal{L}(\hat{x}_t, x_0^{\text{ref}}),
\end{equation}

where $\{\hat{x}_t\}_{t=0}^T$ represents the model-generated trajectory from $x_T$ to $x_0$. 
The objective is to find the trajectory that minimizes $\mathcal{L}$, effectively steering the generated sample toward the target. 
This can be adapted for the denoising stage with either a noise-aware cost function at each timestep $t$ or by estimating $x_0$ to refine the trajectory as needed. 
The gradient update is given by:

\begin{equation} 
x_t \leftarrow x_t - s \cdot \nabla_{x_t} \mathcal{L}(\hat{x}_0, x_0^{ref}),
\label{eq:baseline_control}
\end{equation}

where $s$ is guidance scale. This process requires estimating $\hat{x}_0$, backpropagating gradients through ODESolver ($u_\theta$) to adjust $x_t$, and iteratively refining $x_{t - \Delta t}$. 
Additional details on the baseline algorithm is in the Appendix. 
As it can be observed, this approach depends on accurate $\hat{x}_0$ estimation and substantial computation to ensure that the trajectory remains on the data manifold.

\subsection{Cost Functions}
\begin{table}[!t]
    \centering
    \captionsetup{font=small}
    \scriptsize

    \resizebox{\linewidth}{!}{
    \begin{tabular}{l|cc|ccc}
        \toprule
        \textbf{Method} & \textbf{NFEs} & \textbf{CG Scale}& \textbf{FID} ($\downarrow$) & \textbf{VRAM} ($\downarrow$) & \textbf{Time} ($\downarrow$) \\
        \midrule
        DDIM                 & 50   & -   & 5.39 & 3.67 & 14.22 \\
        MPGD                 & 50   & 1   & 4.24 & 6.56 & 25.01 \\
        MPGD                 & 50   & 10  & 5.46 & 6.56 & 25.01 \\
        \hdashline
        \cellcolor{Apricot!50} \textbf{Ours\nresults{w/ skip grad}}     & 50   & 1   & \textbf{19.28} & 6.56 & 24.95 \\
        \midrule
        RFPP (2-flow)        & 2    & -   & 4.56 & 3.29 & 0.28 \\
        RFPP (2-flow)        & 15   & -   & 4.29 & 3.36 & 2.75 \\
        \hdashline
        \cellcolor{Apricot!50}\textbf{Ours\nresults{w/ backpropagation}} & 15 & 5  & \cellcolor{ForestGreen!30}\textbf{2.77} &  \cellcolor{gray!20}17.98 &  \cellcolor{gray!20}12.79 \\
        \cellcolor{Apricot!50}\textbf{Ours\nresults{w/ skip grad}}    & 15   & 50  & \cellcolor{gray!20}3.13 & \cellcolor{ForestGreen!30}\textbf{6.64} & \cellcolor{ForestGreen!30}\textbf{5.85} \\
        
        \bottomrule
    \end{tabular}
    }
    \caption{Performance of Various guided sampling methods on ImageNet64x64 with 32 batch size inference on A6000 GPU.}
    
    \label{tab:toy_fid}
\end{table}

Notably, explicit $x_0^{ref}$ is unnecessary and can be approximated with appropriate cost functions depending on the downstream tasks. 
Assuming initial Gaussian noise $x_T$ leads to $\hat{x}_0$, the cost function can be defined as:

\begin{equation} 
    \mathcal{L}(\hat{x}_0, x_0^{\text{ref}}) = || \hat{x}_0 - x_0^{\text{ref}} ||_2^2. 
\end{equation}


In inverse problems, let $\mathcal{F}: \mathcal{R}^d \rightarrow \mathcal{R}^n$ represent a degradation operation (e.g., downsampling for super-resolution). 
We then define:

\begin{equation} 
    \mathcal{L}(\hat{x}_0, x_0^{\text{ref}}) = || \mathcal{F}(\hat{x}_0) - x_0^{\text{ref}} ||_2^2. 
    \label{eq:pixel_inverse_loss}
\end{equation}

Here, $x_0^{ref}$ is a degraded sample, and we guide the model to generate $\hat{x}_0$ such that its degraded version matches $x_0^{ref}$. 
For classifier guidance, the cost function can be based on the negative log-likelihood (NLL).
Specifically, given a classifier $p_\phi(c | \hat{x}_0)$, the cost function is:

\begin{equation} 
    \mathcal{L}(\hat{x}_0, c) = - \log p_\phi(c | \hat{x}_0). 
\end{equation}

\noindent\textbf{Remark 1.} Although presented in pixel space, this formulation extends to latent space by introducing a Variational Autoencoder (VAE) encoder ($\mathcal{E}$) and decoder ($\mathcal{D}$).




\section{Proposed Method}
\label{sec:method}

In this section, we introduce our method, \flowchef, which enables free-form control for rectified flow models by presenting an efficient gradient approximation during guided sampling. 
We begin by analyzing the error dynamics of general ordinary differential equations (ODEs) and then explain how the inherent properties of rectified flow models mitigate existing approximation issues. 
Building on these insights, we derive \flowchef, an intuitive yet theoretically grounded approach for free-form controlled image generation applicable to various downstream tasks, including those involving pretrained latent models.

\subsection{Error Dynamics of the ODEs}

Understanding why existing methods often fail and require computationally intensive strategies is crucial. In ODE-based generative models, guiding the sampling process toward a desired target typically involves computing the gradient of a loss function with respect to the model's parameters or state variables. 
As noted in Eq.~(\ref{eq:baseline_control}), even though the denoised output can be estimated using $\hat{x}_0 \leftarrow Sample(x_t, u_\theta(x_t, t))$, backpropagation through the ODE solver is still necessary to obtain $\nabla_{x_t} \mathcal{L}$. 
This raises the question: \textit{Why is backpropagation through the ODE solver necessary?}

Approximating gradient computations is a common approach to reduce computational overhead~\cite{hemanifold, song2023pseudoinverse}. 
However, in models governed by nonlinear ODEs, unregulated gradient approximations can introduce significant errors into the system dynamics. 
This issue is formalized in the following proposition:

\begin{proposition}
\label{prop:error_dynamics}
Let $p_1 \sim \mathcal{N}(0, \mathcal{I})$ be the noise distribution and $p_0$ be the data distribution. 
Let $x_t$ denote an intermediate sample obtained from a predefined forward function $q$ as $x_t = q(x_0, x_1, t)$, where $x_0 \sim p_0$ and $x_1 \sim p_1$.
Define an ODE sampling process $ dx(t) = f(x_t, t) dt$ and quadratic $\mathcal{L} = ||\hat{x}_0 - x_0^{ref}||^2_2$, where $f: \mathcal{R}^d \times [0,T] \rightarrow \mathcal{R}^d$ is an ODESolver.
Then, the error dynamics of ODEs for controlled image generation is governed by:

\begin{equation*}
    \frac{dE(t)}{dt} = -4sE(t) + 2e(t)^T\epsilon(t),
\end{equation*}

where $e(t) = \hat{x}_0 - x_0^{ref}$, $E(t) = e(t)^Te(t)$ is the squared error magnitude, $s > 0$ is the guidance strength, and $\epsilon(t)$ represents the accumulated errors due to non-linearity and trajectory crossovers.

\end{proposition}

The proof of Proposition~\ref{prop:error_dynamics} is provided in the Appendix.
The term $-4sE(t)$ denotes the exponential decay of error due to guidance, while $2e(t)^\top \epsilon(t)$ captures the impact of non-linearity and trajectory crossovers. 
In diffusion models, curved sampling trajectories lead to larger $\epsilon(t)$, hindering convergence. 
In contrast, rectified flow models exhibit straight trajectories with minimal crossovers, causing $\epsilon(t)$ to approach zero and allowing error to decrease exponentially.

To validate our findings, we conduct a toy study comparing classifier guidance on two ODE sampling methods using pretrained IDDPM and Rectified Flow++ (RF++) models on the ImageNet 64x64.
As reported in Table~\ref{tab:toy_fid}, skipping the gradient in DDIM-based sampling increases the FID score, indicating significant $\epsilon(t)$.
Conversely, RF++ converges well and improves the FID score.
These empirical evidences further bolster our hypothesis that Rectified Flow models observe smooth vector field with the help of Proposition~\ref{prop:error_dynamics}.
Although backpropagating through the ODESolver further improves performance, it incurs higher computational costs as highlighted.

\subsection{\flowchef: Steering Within the Vector Field}
\label{sec:flowchef}

Rectified flow models inherently allow error dynamics to converge even with gradient approximations due to their straight-line trajectories and smooth vector fields, as discussed previously. 
Hence, vector field $u_\theta(x_t, t)$ is trained to be smooth, and this smoothness implies that $u_\theta$ changes gradually \textit{w.r.t.} $x_t$.
We formalize our approach with the following assumptions about the Jacobian of the vector field: 

\paragraph{Assumption 1} (Local Linearity):
Within the small neighborhoods around any point $x_t$ along the sampling trajectory, the vector field $u_\theta(x_t, t)$ behaves approximately linearly with respect to $x_t$. 
Doing Taylor series expansion for small perturbations $\delta$, we get:
\begin{equation}
    u_\theta(x_t + \delta, t) \approx u_\theta(x_t, t) + J_{u_\theta}(x_t, t)\delta,
\end{equation}

where $J_{u_\theta}(x_t, t) = \frac{du_\theta(x_t, t)}{dx_t}$ is the Jacobian matrix of $u_\theta$ with respect to $x_t$.

\paragraph{Assumption 2} (Constancy of the Jacobian):
The Jacobian $J_{u_\theta}(x_t, t)$ varies slowly with respect to $x_t$ within these small neighborhoods. 
Therefore, for small $\delta$, it can be approximated as constant:
\begin{equation}
    J_{u_\theta}(x_t + \delta, t) \approx J_{u_\theta}(x_t, t).
\end{equation}

Under these assumptions, we derive the following gradient relationship between $\nabla_{x_t} \mathcal{L}$ and $\nabla_{\hat{x}_0} \mathcal{L}$:

\begin{figure*}[t]
    \centering
    \includegraphics[width=\linewidth]{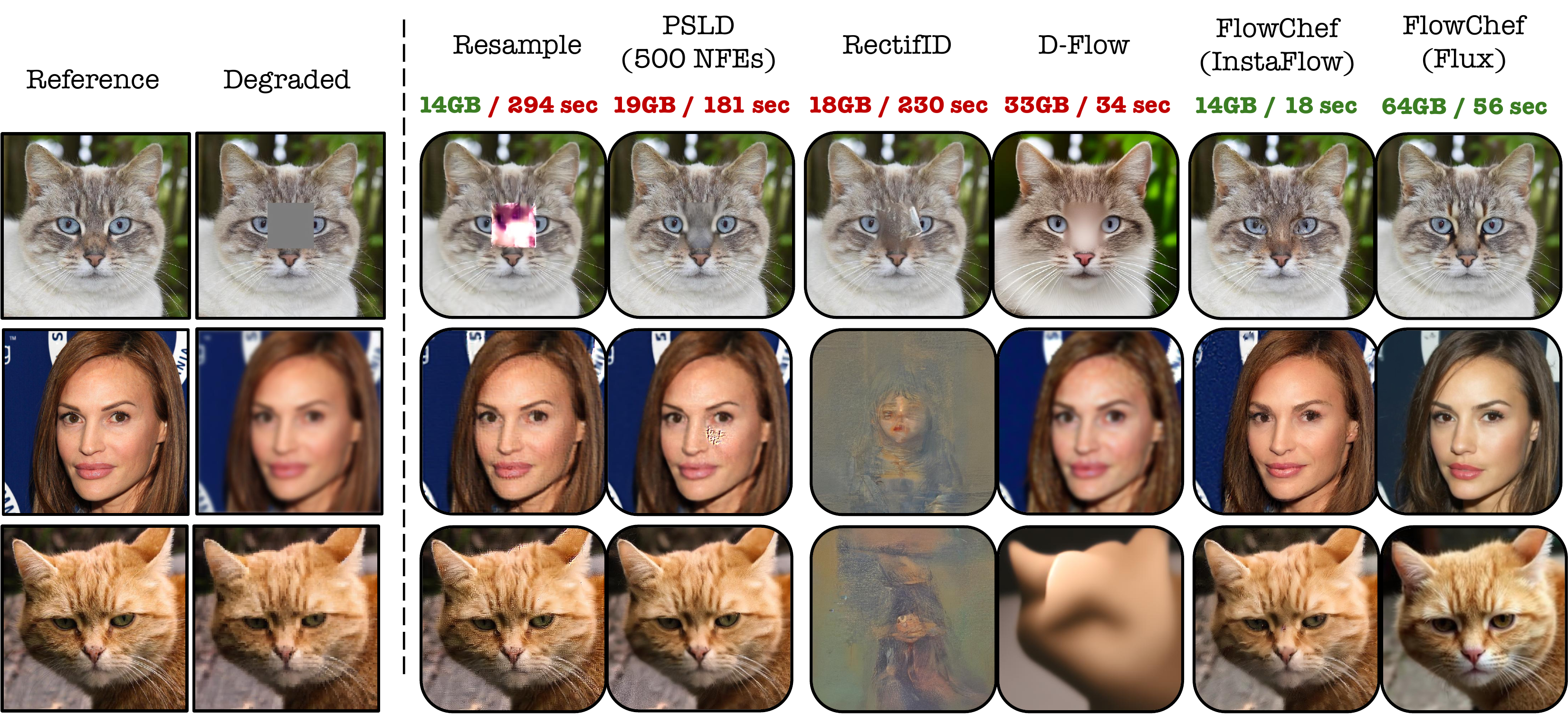}
    \caption{\textbf{Qualitative results on linear inverse problems.} All baselines are implemented on stable diffusion v1.5, except \flowchef~Flux variant. Results are reported for VRAM and time on an A100 GPU at 512 x 512 resolution, with Flux experiments at 1024 x 1024. Best viewed when zoomed in.}
    \label{fig:ldm_ip}
\end{figure*}

\begin{lemma}[Gradient Relationship]
    Let $u_\theta : \mathcal{R}^d \times [0,T] \rightarrow \mathcal{R}^d$ be the velocity function with the parameter $\theta$.
    Then the gradient of the cost function ($\nabla_{x_t} \mathcal{L}$) at any timestep $t$ can be approximated as:
    \begin{equation}
        \nabla_{x_t} \mathcal{L} = (I + t \cdot J_{u_\theta})^T \nabla_{\hat{x}_0} \mathcal{L}. 
    \end{equation}
    \label{lemma:grad_approx}
\end{lemma}

Therefore, we get $\epsilon(t) = t\cdot J_{u_\theta}(x_t, t)^T \nabla_{\hat{x}_0}\mathcal{L}$.
Importantly, when either $t \rightarrow 0$ or $J_{u_\theta}$ varies slowly (Assumption 2), the matrices $I + t \cdot J_{u_\theta}(x_t, t)$ are close to the identity matrix.
We further provide empirical evidence about this on pretrained rectified flow models by analyzing the gradients and convergence \textit{w.r.t.} denoising steps in the Appendix.
Where we observe that gradient direction improves linearly and quickly converges to $x_0^{ref}$ as $t \rightarrow 0$.
Under this approximation, the difference between the two error dynamics becomes negligible.
Since the $\epsilon(t)$ introduces only a small correction, it leads to the convergence in error dynamics as $t \rightarrow 0$.
Combining the results of Preposition~\ref{prop:error_dynamics}, Assumption 1 and 2, and Lemma~\ref{lemma:grad_approx}, we obtain the following theorem with straightforward proof that facilitates the controlled generation for rectified flow models in the most computationally efficient way:

\begin{theorem}(Informal)
    Given the above assumption and notations, the update rule for the vector field driven by $u_\theta$ for the free-form controlled generation is:
    \begin{equation}
        x_{t - \Delta t} = x_t + \Delta t \cdot u_\theta(x_t, t) - s' \nabla_{\hat{x}_0}\mathcal{L},
        \label{eq:theorem}
    \end{equation}
    where $s'$ is the guidance scale.
    \label{theorem:method}
\end{theorem}

The formal statement and proof are provided in the Appendix. 
This theorem forms the core of \flowchef, enabling controlled generation efficiently.

\begin{algorithm}[!t]
\DontPrintSemicolon
\textbf{Input:} Pretrained Rectified-flow model $u_\theta$, input noise sample $x_T\sim N(0, I)$, target data sample $x_0^{ref}$, and $\mathcal{L}$ cost function. \;

\For {$t \in \{T...0\}$}
{
 $v \leftarrow u_\theta(x_t, t)$ \;
 $dt \leftarrow 1/T$ \;
 $x_t \leftarrow x_t.require\_grad\_(True)$ \;
 
  \For {N \textnormal{steps}}
    {  
    $\hat{x}_0 \leftarrow x_t + t\cdot v$
    
    $loss \leftarrow \mathcal{L}(\hat{x}_0, 
    x_0^{ref})$
    
    $x_t \leftarrow$ Optimize($x_t$, loss) \quad \quad // Lemma~\ref{lemma:grad_approx}
    } 

    $x_{t-1} \leftarrow x_t + dt\cdot v$ \quad \quad \quad \quad \quad // Theorem~\ref{theorem:method}

}

\textbf{RETURN} $x_0$
\caption{Proposed \flowchef~(generalized).}
\label{algo:algo1}
\end{algorithm}

\paragraph{Algorithm Overview.}
Algorithm~\ref{algo:algo1}  provides a generalized overview of \flowchef. 
A key feature of \flowchef\ is that it starts from any random noise $x_T \sim \mathcal{N}(0, I)$ and still converges to the desired distribution or sample without inversion. 
At each timestep $t$, we first estimate the $\hat{x}_0$.
Then we calculate the loss $\mathcal{L}(\hat{x}_0, x_0^{\text{ref}})$.
At last, we directly optimize $x_t$ using the gradient $\nabla_{\hat{x}_0} \mathcal{L}$, as per Lemma~\ref{lemma:grad_approx}.
\textbf{That's all we need!}
We may repeat this optimization $N$ times per denoising step to stabilize gradients and improve convergence, though we found $N=1$ sufficient in most cases. 
Important hyperparameters include the learning rate and total number of function evaluations (NFEs) $T$. 
Selecting optimal values for $T$ and the learning rate is crucial to maintain gradients within a suitable range, uphold Jacobian constancy (Assumption 2), and avoid adversarial effects.
To illustrate this, we analyze the effects of total \flowchef\ guidance steps on the Flux model (see Figure~\ref{fig:algo_hyperparameter}). 
Detailed study on this is in Appendix.



\section{Experiments}
\label{sec:expeirments}

\begin{table*}[ht]
    \centering
    \captionsetup{font=small}
    \scriptsize

    \resizebox{\linewidth}{!}{%

    \begin{tabular}{l ccc ccc cccc}
        \toprule
        \multirow{2}{*}{\textbf{Method}} & \multicolumn{3}{c}{\textbf{BoxInpaint}} & \multicolumn{3}{c}{\textbf{Deblurring}} & \multicolumn{3}{c}{\textbf{Super Resolution}} \\
        \cmidrule(lr){2-4} \cmidrule(lr){5-7} \cmidrule(lr){8-10}
        & \textbf{PSNR} ($\uparrow$) & \textbf{SSIM} ($\uparrow$) & \textbf{LPIPS} ($\downarrow$) & \textbf{PSNR} ($\uparrow$) & \textbf{SSIM} ($\uparrow$) & \textbf{LPIPS} ($\downarrow$) & \textbf{PSNR} ($\uparrow$) & \textbf{SSIM} ($\uparrow$) & \textbf{LPIPS} ($\downarrow$) \\
        \midrule
        \textsuperscript{{\color{BrickRed}\textbf{Easy Scenarios}}} &  \\[-3pt]
        Degraded & 21.79 & 74.76 & 10.92 & 20.17 & 54.03 & 22.20 & 24.68 & 77.57 & 11.67 \\
        OT-ODE       & 19.11 & \cellcolor{gray!20}77.86 & 13.49 & 21.86 & 62.51 & 15.14 & 21.64 & 62.23 & 26.64 \\
        PnP-Flow      & 22.12 & 68.02 & 14.70 & 22.00 & 65.79 & 15.95 & \cellcolor{gray!20}22.42 & 68.06 & 14.91 \\
        D-FLow        & 20.37 & 70.06 & 13.67 & 20.22 & 61.99 & 14.51 & 21.60 & \cellcolor{gray!20}69.89 & 12.29 \\
        FreeDoM &20.87 &74.79 &13.92 &20.21 &69.73 &13.22 &21.15 &77.54 &12.12 \\
        DPS &\cellcolor{gray!20}23.61 &74.79 &\cellcolor{gray!20}9.35 &\cellcolor{gray!20}22.49 &\cellcolor{gray!20}69.73 &\cellcolor{gray!20}10.23 &23.94 &77.54 &\cellcolor{gray!20}8.46 \\

        \hdashline
        \cellcolor{Apricot!50} \textbf{\flowchef~(ours)}          & \cellcolor{ForestGreen!30}\textbf{26.32} & \cellcolor{ForestGreen!30}\textbf{87.70} & \cellcolor{ForestGreen!30}\textbf{3.36} & \cellcolor{ForestGreen!30}\textbf{27.69} & \cellcolor{ForestGreen!30}\textbf{86.43} & \cellcolor{ForestGreen!30}\textbf{2.66} & \cellcolor{ForestGreen!30}\textbf{26.00} & \cellcolor{ForestGreen!30}\textbf{80.15} & \cellcolor{ForestGreen!30}\textbf{4.43} \\

        \midrule
        \textsuperscript{{\color{BrickRed}\textbf{Hard Scenarios}}} &  \\[-3pt]
        Degraded &  18.75 & 65.12 & 22.54 & 16.83 & 30.02 & 54.04 & 20.77 & 55.85 & 38.16 \\
        OT-ODE        & 16.37 & \cellcolor{gray!20}67.35 & 19.22 & 17.89 & 34.02 & 29.68 & 18.19 & 39.43 & 36.84 \\
        PnP-Flow      & 20.44 & 61.96 & 17.53 & \cellcolor{gray!20}19.50 & \cellcolor{gray!20}50.54 & 22.00 & 21.35 & \cellcolor{gray!20}61.78 & 17.78 \\
        D-FLow        & 18.34 & 62.62 & 19.94 & 16.93 & 34.13 & 25.31 & 20.01 & 56.46 & 17.64 \\

        FreeDoM &18.88 &65.07 &16.83 &16.50 &34.88 &18.91 &19.58 &55.84 &14.12 \\
        DPS &\cellcolor{gray!20}20.68 &65.06 &\cellcolor{gray!20}13.06 &17.58 &34.89 &\cellcolor{gray!20}15.86 &\cellcolor{gray!20}21.52 &55.90 &\cellcolor{gray!20}10.31 \\

        \hdashline
        \cellcolor{Apricot!50}\textbf{\flowchef~(ours)}         & \cellcolor{ForestGreen!30}\textbf{21.45} & \cellcolor{ForestGreen!30}\textbf{78.75} & \cellcolor{ForestGreen!30}\textbf{7.73} & \cellcolor{ForestGreen!30}\textbf{20.31} & \cellcolor{ForestGreen!30}\textbf{52.73} & \cellcolor{ForestGreen!30}\textbf{10.64} & \cellcolor{ForestGreen!30}\textbf{21.62} & \cellcolor{ForestGreen!30}\textbf{60.33} & \cellcolor{ForestGreen!30}\textbf{10.18} \\
        \bottomrule
    \end{tabular}
    }
    \caption{Pixel-space model-based evaluations for tackling the linear inverse problems. SSIM \& LPIPS results are multiplied by 100.}
        \label{tab:pixel_ip_exntended}

\end{table*}

\begin{table*}[ht]
    \centering
    \captionsetup{font=small}
    \scriptsize

    \resizebox{\linewidth}{!}{%

    \begin{tabular}{lccccccccc}
        \toprule
        \multirow{2}{*}{\textbf{Method}} & \multicolumn{3}{c}{\textbf{BoxInpaint}} & \multicolumn{3}{c}{\textbf{Super Resolution}} & \multicolumn{3}{c}{\textbf{Deblurring}} \\
        \cmidrule(lr){2-4} \cmidrule(lr){5-7} \cmidrule(lr){8-10}
        & \textbf{PSNR} ($\uparrow$) & \textbf{SSIM} ($\uparrow$) & \textbf{LPIPS} ($\downarrow$) & \textbf{PSNR} ($\uparrow$) & \textbf{SSIM} ($\uparrow$) & \textbf{LPIPS} ($\downarrow$) & \textbf{PSNR} ($\uparrow$) & \textbf{SSIM} ($\uparrow$) & \textbf{LPIPS} ($\downarrow$) \\
        \midrule
        \textsuperscript{{\color{BrickRed}\textbf{Diffusion based methods}}} &  \\[-3pt]
        Resample      & 20.12 & 79.94 & 19.36 & \cellcolor{ForestGreen!30}26.91 & \cellcolor{ForestGreen!30}70.91 & \cellcolor{ForestGreen!30}30.75 & \cellcolor{gray!20}25.27 & \cellcolor{gray!20}62.97 & \cellcolor{ForestGreen!30}41.94 \\
        PSLD (500 NFEs) & \cellcolor{ForestGreen!30}28.30 & \cellcolor{ForestGreen!30}93.81 & \cellcolor{ForestGreen!30}4.49 & 25.79  & \cellcolor{gray!20}65.15 & \cellcolor{gray!20}33.27 & \cellcolor{ForestGreen!30}26.64 & \cellcolor{ForestGreen!30}65.44 & 43.10 \\
        PSLD (100 NFEs) & \cellcolor{gray!20}26.90 & \cellcolor{gray!20}93.13 & \cellcolor{gray!20}5.29 & 21.95  & 54.67 & 46.08 & 21.25 & 51.62 & 51.92 \\
        
        \midrule
        
        \textsuperscript{{\color{BrickRed}\textbf{Flow based methods}}} &  \\[-3pt]
        D-Flow        & 19.68  & 65.01 & 27.79 & 20.23 & 60.55 & 50.30 & \cellcolor{gray!20}22.42 & \cellcolor{ForestGreen!30}64.43 & \cellcolor{gray!20}53.04 \\
        RectifID      & \cellcolor{gray!20}23.81  & \cellcolor{gray!20}75.13 & 10.50 & 10.36 & 31.55 & 67.08 & 10.40      & 31.16      & 66.60 \\
        
        \hdashline
        
        \cellcolor{Apricot!50}\textbf{\flowchef~(InstaFlow)}          & 22.94 & 73.55 & \cellcolor{gray!20}\textbf{9.94} & \cellcolor{ForestGreen!30}\textbf{25.83} & \cellcolor{ForestGreen!30}\textbf{64.73} & \cellcolor{ForestGreen!30}\textbf{31.38} & \cellcolor{ForestGreen!30}\textbf{22.50} & 47.42 & \cellcolor{ForestGreen!30}\textbf{42.54} \\
        
        \cellcolor{Apricot!50}\textbf{\flowchef~(Flux)} & \cellcolor{ForestGreen!30}\textbf{25.74} & \cellcolor{ForestGreen!30}\textbf{82.99} & \cellcolor{ForestGreen!30}\textbf{9.40} & \cellcolor{gray!20}\textbf{20.25} & \cellcolor{gray!20}\textbf{64.34} & \cellcolor{gray!20}\textbf{41.88} & 18.98 & \cellcolor{gray!20}\textbf{64.37} & 53.43 \\

        \bottomrule
        
    \end{tabular}
    }
     \caption{Latent-space model based evaluations for tackling the linear inverse problems. SSIM \& LPIPS results are multiplied by 100.}
         \label{tab:latent_ip}

\end{table*}

We evaluate \flowchef~across multiple tasks: 
(1) Linear inversion problems on pixel- and latent-space models, 
(2) Image editing, and 
(3) Classifier-guided style transfer. 
Overall, \flowchef~demonstrates superior performance across all tasks, significantly reducing compute and time costs compared to baselines. 
Notably, \flowchef~extends seamlessly to image editing tasks without inversion or additional memory overhead, allowing it to operate on recent SOTA T2I models, such as Flux, without encountering out-of-memory (OOM) errors.

\subsection{Linear Inversion Problems}

We evaluate \flowchef~against several baselines on three common linear tasks: box inpainting, super-resolution, and Gaussian deblurring, under varying difficulty levels. 
We extend both \flowchef~and the baselines to latent-space models to simulate real-world applications, reporting results on PSNR, SSIM~\cite{ssim}, and LPIPS~\cite{zhang2018unreasonable} across 200 images from CelebA~\cite{liu2015faceattributes} and AFHQ-Cat~\cite{Choi_2020_CVPR}.
Memory requirements and computation time are also analyzed. 

\subsubsection{Pixel-space models}

\begin{figure*}[t]
    \centering
    \includegraphics[width=\linewidth]{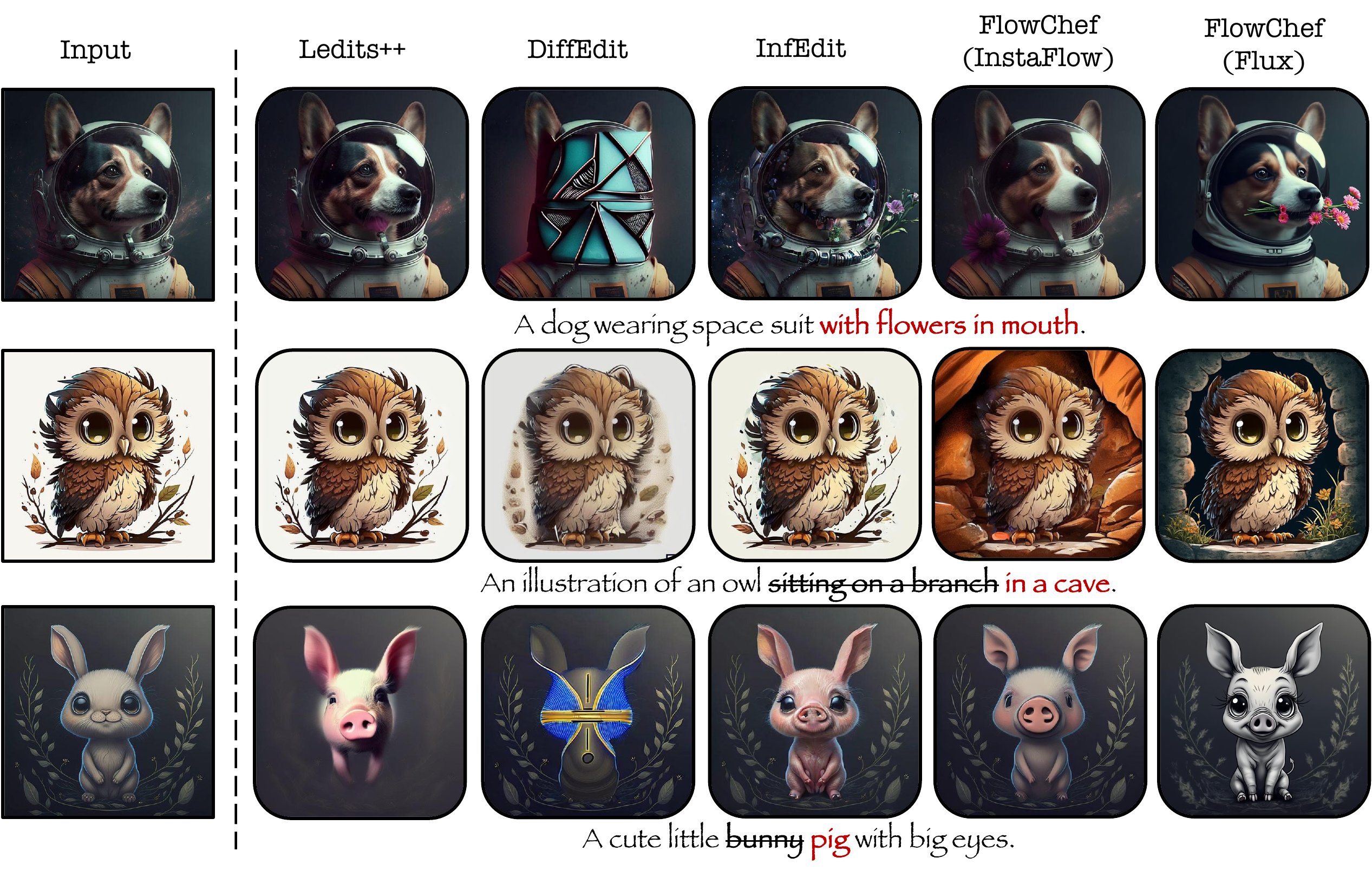}
    \caption{\textbf{Qualitative results on image editing.} As illustrated, our method attains the SOTA performance on comparison inversion-free methods. While \flowchef~(Flux) variant achieves better quality and edits.}
    \label{fig:editing}
\end{figure*}

As \flowchef~requires straightness and no crossovers, we select the Rectified-Flow++ pretrained models~\cite{lee2024improving}.
We compare \flowchef~with recent flow-based methods OT-ODE~\cite{pokle2024trainingfree}, D-Flow~\cite{ben-hamu2024dflow}, and PnP-Flow (concurrent work)~\cite{martin2024pnp}, implementing the former two baselines manually due to lack of open-source access and tuning them for optimal performance. 
Additionally, we extend two diffusion-based baselines, DPS~\cite{chung2022diffusion} and FreeDoM~\cite{yu2023freedom}, for the RFMs.
For comparisons, we use the Rectified-Flow++ models that are pretrained on FFHQ (for CelebA) and AFHQ-Cat datasets. 
Experiments are conducted for 64x64 image resolutions.
Hyper-parameters for each method are reported in the Appendix.
Our selected tasks include: (1) Box inpainting with 20x20 and 30x30 centered masks, (2) Super-resolution with 2x and 4x scaling factors, and (3) Gaussian deblurring with an 11x11 kernel at intensities of 1.0 and 10.0, with added Gaussian noise at $\sigma=0.05$ for robustness.

\paragraph{Results.}
We present the quantitative and qualitative evaluation results in Table~\ref{tab:pixel_ip_exntended} and Appendix, respectively.
It can be observed that \flowchef~significantly improves the performance on both easy and hard settings across the tasks and all metrics consistently. 
Notably from Table~\ref{tab:vram_ip}, we find that the \flowchef~is also the fastest and most memory efficient. 
Surprisingly, diffusion-based extended baseline (DPS) significantly outperforms even recent baselines.
However, DPS requires backpropagation through billions of parameters of ODESolver. 
While the concurrent gradient-free work, PnP-Flow, outperforms many other baselines, \flowchef~leads the benchmark.

\subsubsection{Latent-space models.}
\begin{table}[!t]
    \centering
    \captionsetup{font=small}
    \scriptsize

    \resizebox{\linewidth}{!}{
    \begin{tabular}{l cccc}
        \toprule
        \textbf{Metric} & \textbf{OT-ODE} & \textbf{PnP-Flow} & \textbf{D-Flow} & \textbf{\flowchef}\\
        \midrule
        \textbf{VRAM (GB)} & 0.70 & \cellcolor{ForestGreen!30}0.40 & 6.44 & \cellcolor{gray!20}\textbf{0.43}\\
        \textbf{Time (sec)} & 10.39 & \cellcolor{gray!20}5.23 & 80.42 & \cellcolor{ForestGreen!30}\textbf{4.31}\\
        \bottomrule
    \end{tabular}
    }
    \caption{Compute requirement comparisons on a A6000 GPU.}
    \label{tab:vram_ip}
    \vspace{-0.5cm}
\end{table} 

Flow-based baselines are not extended to the latent space models as either they are already very computationally heavy or require extra Jacobian calculations to support the non-linearity introduced by the VAE models.
We adapt D-Flow~\cite{ben-hamu2024dflow} and RectifID~\cite{sun2024rectifid} as flow-based baselines, adding diffusion-based baselines PSLD-LDM~\cite{rout2024solving} and Resample~\cite{songsolving} for comparison.
We use InstaFlow~\cite{liu2023instaflow} (Stable Diffusion v1.5 variant) and Flux models as a baseline for flow-based approaches and utilize the original Stable Diffusion v1.5 checkpoint for the diffusion-based baselines.
We perform all tasks in 512 x 512 resolution, increasing to 1024 x 1024 for Flux experiments.
Our task settings are: (1) Box inpainting with a 128x128 mask, (2) Super-resolution at 4x scaling, and (3) Gaussian deblurring with a 50x50 kernel at intensity 5.0, all without extra Gaussian noise. 
For consistency, settings are doubled for Flux to a 256x256 mask, 8x super-resolution scaling, and 10.0 deblurring intensity.
As VAE encoders add extra unwanted nonlinearity, pixel-level cost functions alone may not be optimal.
Hence, we calculate the loss in the latent space only for the box inpainting task (as the degradation function is known with $\sigma=0$), allowing us to extend to image editing later. 
For super-resolution and deblurring, we stick with the pixel-level cost functions.
We further detail the task-specific settings and hyperparameters in the Appendix.

\begin{figure}[t]
    \centering
    \includegraphics[width=\linewidth]{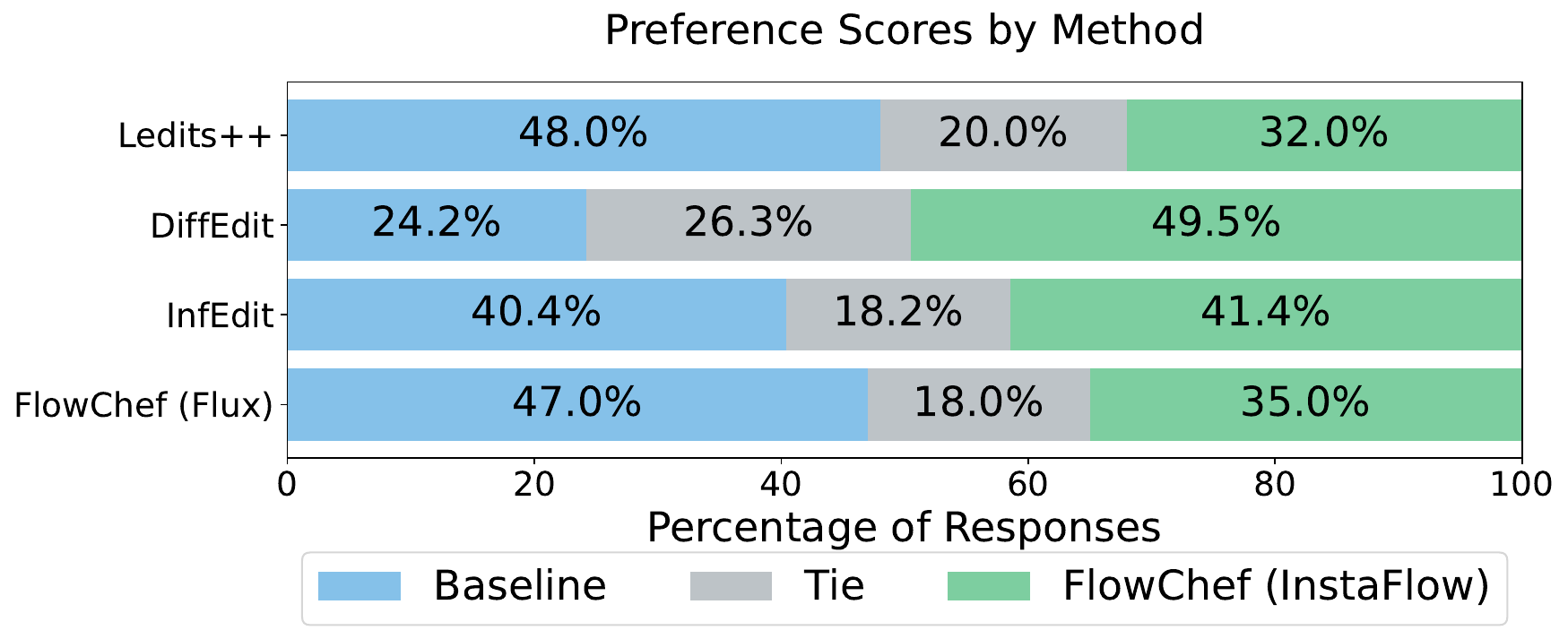}
    \caption{Human preference analysis for image editing.}
    \label{fig:human_preference}
\end{figure}

\paragraph{Results.}
Quantitative and qualitative results in Figure~\ref{fig:ldm_ip} and Table~\ref{tab:latent_ip} show that \flowchef~achieves SOTA performance for flow-based methods.
However, a huge gap still remains \textit{w.r.t.} the diffusion-based methods like Resample and PSLD.
Notably, these baselines take about 5 minutes and 3 minutes, respectively, per image (see Figure~\ref{fig:ldm_ip}), while \flowchef~only takes only 18 seconds and less memory (only 14GB).
None of the existing flow-based methods can be extended to Flux due to memory constraints.
But \flowchef~can seamlessly be applied, which further improves the performance.
We find that \flowchef~(Flux) reduces the artifacts in the images completely but observes the slight degradation in color dynamics.
We attribute this to the observed nonlinearity in the trajectory of Flux (detailed discussion in Appendix).

\subsection{Image Editing}

We extend \flowchef~for image editing on Flux and InstaFlow models, with Algorithm 2 detailing the implementation. 
This extension reduces \flowchef’s sensitivity to hyper-parameters. 
Currently, the approach requires a user-provided mask for controlled editing but can be expanded to attention-based techniques.
Therefore, we select the baselines that also accept the user-provided mask for holistic comparisons.
Due to their optimization constraints, existing baselines for classifier guidance cannot be applied to image editing. 
For comparison, we use diffusion-based SOTA methods Ledits++~\cite{brack2024ledits++} (which requires the inversion), DiffEdit~\cite{couairon2023diffedit} and InfEdit~\cite{xu2023inversion}, alongside RF-Inversion~\cite{rout2024semantic} (the only concurrent flow-based editing framework).
We perform large-scale evaluations on PIE-Bench~\cite{ju2023direct}.
For fair comparisons, we use PIE-Bench-provided ground truth masks for controlling all editing methods.
Additionally, we provide preliminary comparisons with RF-Inversion for ``wearing glasses'' on randomly selected SFHQ faces~\cite{david_beniaguev_2022_SFHQ}.

\paragraph{Results.}
A human preference evaluation on randomly selected 100 PIE-Bench edits (see Figure~\ref{fig:human_preference}) shows  \flowchef~(InstaFlow) outperforming DiffEdit and competing with InfEdit. 
Although Ledits++ scored highest, it requires inversion, resulting in higher VRAM and time requirements. 
Importantly, \flowchef~on Flux achieves performance comparable to Ledits++ without inversion.
Comparisons with RF-Inversion show that \flowchef~reduces time by almost 50\% without needing inversion and achieves competitive performance, with additional detailed quantitative and qualitative results in the Appendix.

\subsection{Classifier Guidance: Style Transfer}

\begin{figure}[t]
    \centering
    \includegraphics[width=\linewidth]{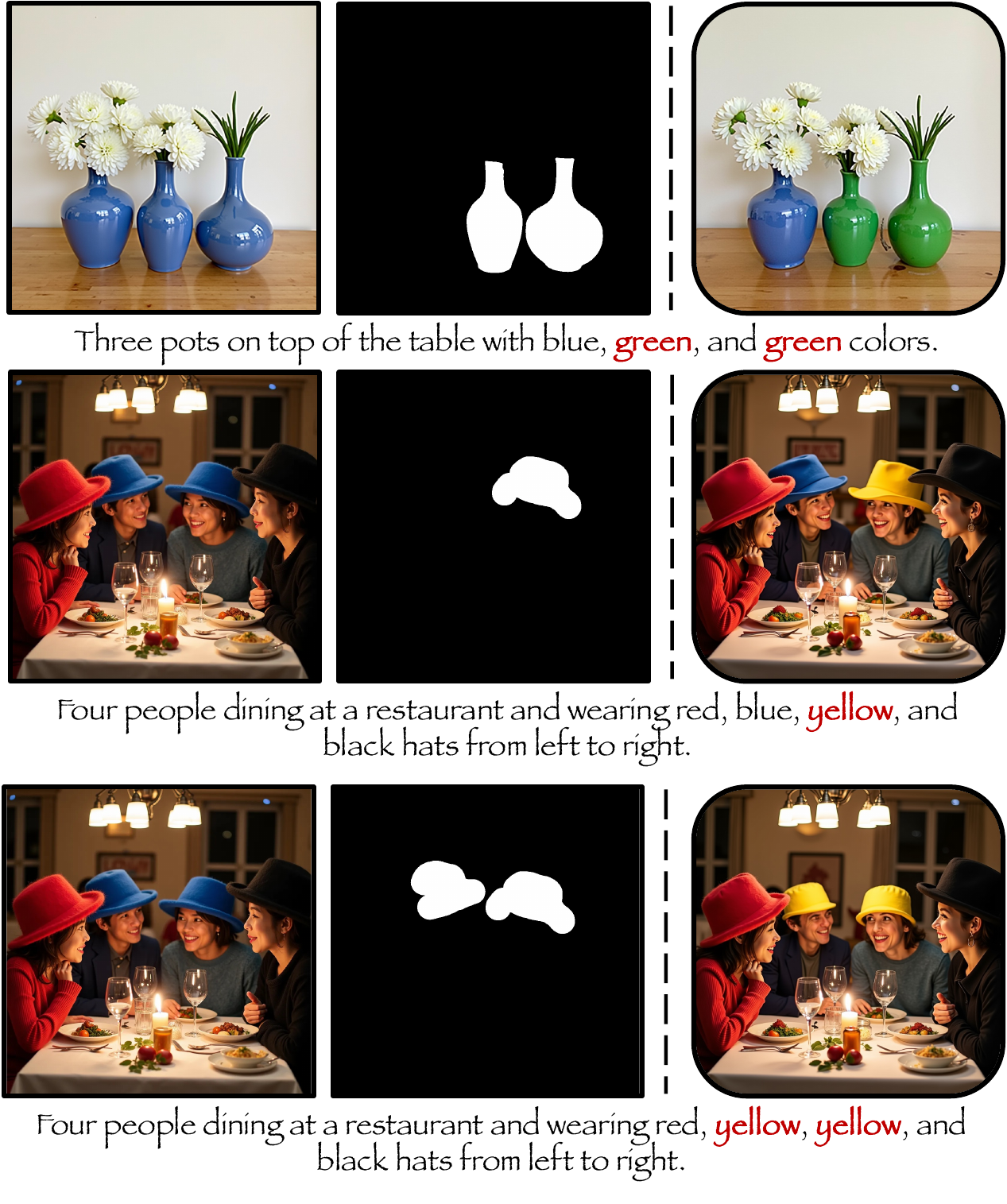}
    \caption{\flowchef~(Flux) multi object editing examples.}
    \label{fig:appendix_multiedit}
\end{figure}

\begin{figure}[t]
    \centering
    \includegraphics[width=\linewidth]{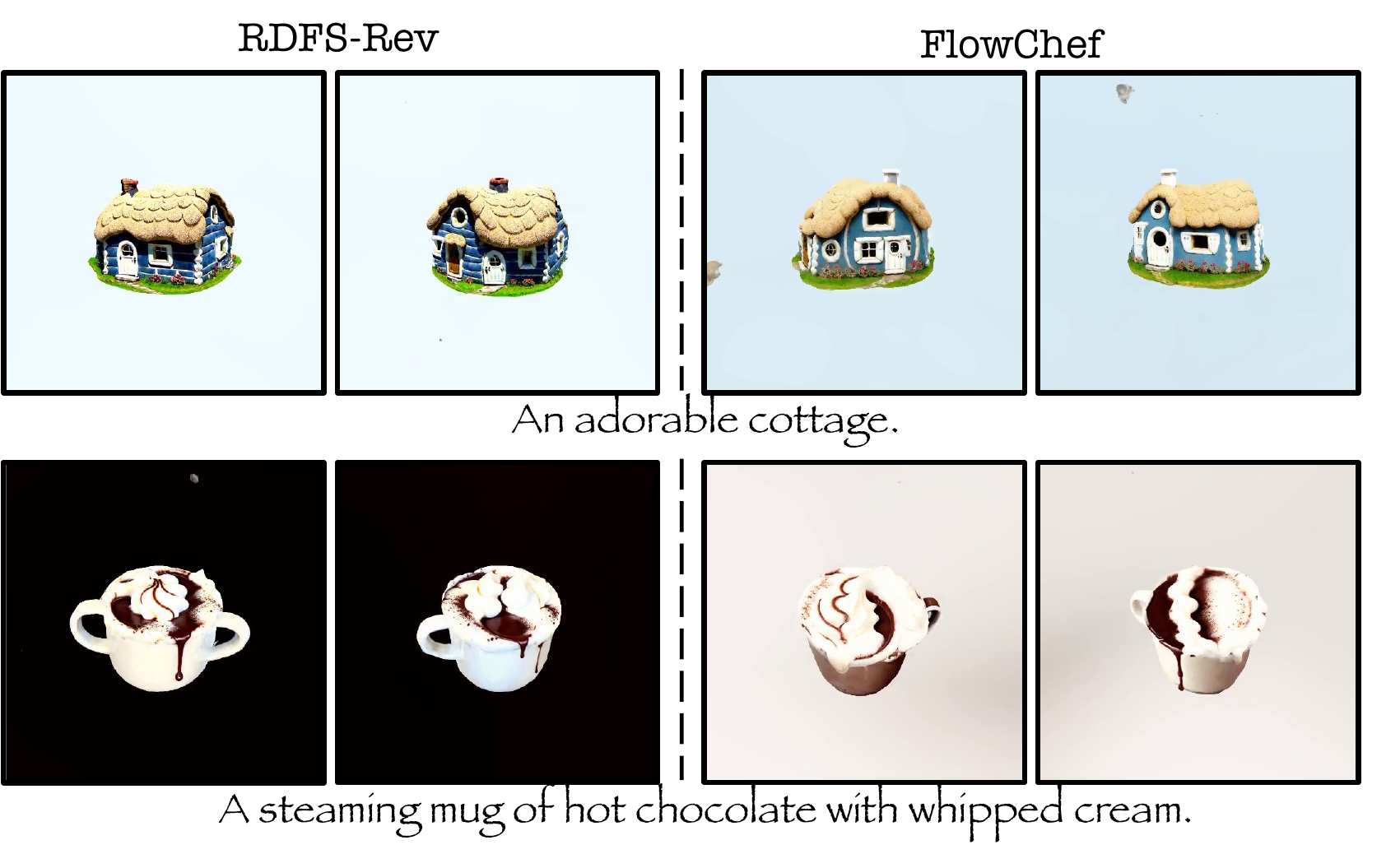}
    \caption{Extending \flowchef~to 3D multiview synthesis.}
    \label{fig:appendix_3d_views}
\end{figure}

\begin{table}[!t]
    \centering
    \captionsetup{font=small}
    \scriptsize

    \resizebox{\linewidth}{!}{
    \begin{tabular}{l|cc|cc}
        \toprule
        \textbf{Method} & \textbf{CLIP-I} ($\uparrow$) & \textbf{CLIP-T} ($\uparrow$) & \textbf{VRAM} & \textbf{Time} \\
        \midrule
        FreeDoM        & \cellcolor{gray!20}\textbf{0.5343} & 0.2541 & 17GB & 80 sec \\
        MPGD           & 0.5285 & \cellcolor{gray!20}\textbf{0.2616} & \cellcolor{gray!20}\textbf{16GB} & 20 sec \\
        
        RetifID        & 0.4583 & 0.1702 & 18GB & 30 sec \\
        D-Flow         & 0.4851 & 0.2591 & 23GB & 5 sec \\
        \hdashline
        \cellcolor{Apricot!50} \textbf{\flowchef\nresults{(10 NFEs)}}      & 0.5044 & \cellcolor{ForestGreen!30}\textbf{0.2655} & \cellcolor{ForestGreen!30} & \cellcolor{ForestGreen!30}\textbf{2 sec} \\
        \cellcolor{Apricot!50} \textbf{\flowchef\nresults{(30 NFEs)}}       & 0.5301 & 0.2600 & \cellcolor{ForestGreen!30} & \cellcolor{gray!20}\textbf{7 sec} \\
        \cellcolor{Apricot!50} \textbf{\flowchef\nresults{(30 NFEs $\times$ 2)}}      & \cellcolor{ForestGreen!30}\textbf{0.5531} & 0.2478 & \cellcolor{ForestGreen!30}\multirow{-3}{*}{\textbf{14GB}} & 12 sec \\
        \bottomrule
    \end{tabular}
    }
    \caption{Comparison of Various Classifier Guided Style Transfer.}
    \label{tab:style_transfer}
    
\end{table}

We conducted classifier-guided style transfer experiments using 100 randomly selected style reference images paired with 100 random prompts. The objective was to generate stylistic images that align visually with the reference style while adhering to the prompt. 
A pretrained CLIP model was used for evaluation, and we report both CLIP-T and CLIP-S scores~\cite{Radford2021LearningTV}. For baseline comparisons, we included diffusion-based methods FreeDoM and MPGD and flow-based methods D-Flow and RectifID, which were extended for this task. 
The backbone was fixed to Stable Diffusion v1.5 (SDv1.5), with \flowchef~evaluated in its InstaFlow variant to ensure a consistent comparison. Both quantitative and qualitative results are presented in Table~\ref{tab:style_transfer}, demonstrating the effectiveness of \flowchef~in this setup.

\subsection{Extended Applications}
\label{sec:appendix_coolstuff}

To highlight the versatility and effectiveness of \flowchef, we extended our method to tackle multi-object image editing and 3D multiview generation.
Figure~\ref{fig:appendix_multiedit} demonstrates \flowchef~(Flux) performing complex multi-object edits, such as simultaneously modifying two pots and hats. Notably, this capability relies on the base model's ability to understand textual instructions effectively. \flowchef~leverages this strength of Flux, achieving edits without requiring inversion, a significant advantage over traditional methods.
In Figure~\ref{fig:appendix_3d_views}, we explore \flowchef's multiview synthesis capability, inspired by Score Distillation Sampling (SDS)~\cite{poole2023dreamfusion}. By incorporating the core idea of \flowchef~for model steering into recent work on RFDS~\cite{yang2024text}, we evaluate its effectiveness for 3D view generation. While \flowchef~does not improve inference efficiency or reduce cost compared to RFDS-Rev~\cite{yang2024text}, it demonstrates competitive performance in generating high-quality multiview outputs. 
These results underline the adaptability of \flowchef, showcasing its potential for advanced generative tasks such as multi-object editing and 3D synthesis, while maintaining the state-of-the-art quality expected from RFMs.
\section{Conclusion}

In this work, we introduced \flowchef, a versatile flow-based approach that unifies key tasks in controlled image generation, including linear inverse problems, image editing, and classifier-guided style transfer. Extensive experiments show that \flowchef~outperforms baselines across all tasks, achieving state-of-the-art performance with reduced computational cost and memory usage. Notably, \flowchef~enables inversion-free editing and scales to SOTA T2I models like Flux without memory issues. Our results demonstrate \flowchef’s adaptability and efficiency, offering a unified solution for both pixel and latent spaces across diverse architectures and practical constraints.

\section*{Acknowledgments}
MP and YY are supported by NSF RI grants \#1750082 and \#2132724. 
We thank the Research Computing (RC) at Arizona State University (ASU) and \href{https://www.cr8dl.ai/}{cr8dl.ai} for their generous support in providing computing resources.
The views and opinions of the authors expressed herein do not necessarily state or reflect those of the funding agencies and employers.

{
    \small
    \bibliographystyle{ieeenat_fullname}
    \bibliography{main}
}



\clearpage
\setcounter{page}{1}
\maketitlesupplementary

\section{Supplementary Overview}
This supplementary material contains proofs, detailed results, discussion, and qualitative results:

\begin{itemize}
    \item Section~\ref{sec:appendix_proposotion}: Proposition~\ref{prop:error_dynamics} proof.
    \item Section~\ref{sec:appendix_theorem}: Theorem~\ref{theorem:method} proof.
    \item Section~\ref{sec:appendix_numerical_accuracy}: Numerical accuracy analysis.
    \item Section~\ref{sec:appendix_related}: Extended related works.
    \item Section~\ref{sec:appendix_findings}: Empirical study of pixel and latent models.
    \item Section~\ref{sec:appendix_algos}: Detailed algorithms.
    \item Section~\ref{sec:appendix_setup}: Experimental setup details.
    \item Section~\ref{sec:appendix_rfinversion}: RF-Inversion \textit{vs.} \flowchef.
    \item Section~\ref{sec:appendix_hyperparams}: Hyperparameter study.
    \item Section~\ref{sec:appendix_qualitatives}: Qualitative Results.
    \item Section~\ref{sec:appendix_futurework}: Limitations \& Future Work
\end{itemize}

\section{Proof of the Proposition}
\label{sec:appendix_proposotion}

\paragraph{Proposition 4.1.}
Let $p_1 \sim \mathcal{N}(0, \mathcal{I})$ be the noise distribution and $p_0$ be the data distribution. 
Let $x_t$ denote an intermediate sample obtained from a predefined forward function $q$ as $x_t = q(x_0, x_1, t)$, where $x_0 \sim p_0$ and $x_1 \sim p_1$.
Define a ODE sampling process $ dx(t) = f(x_t, t) dt$ and quadratic $\mathcal{L} = ||\hat{x}_0 - x_0^{ref}||^2_2$, where $f: \mathcal{R}^d \times [0,T] \rightarrow \mathcal{R}^d$ is a nonlinear function parameterized by $\theta$.
Then, under Assumption 1, the error dynamics of ODEs for controlled image generation are governed by:

\begin{equation*}
    \frac{dE(t)}{dt} = -4sE(t) + 2e(t)^T\epsilon(t),
\end{equation*}

where $e(t)$ is $\hat{x}_0 - x_0^{ref}$, $E(t) = e(t)^Te(t)$ is the squared error magnitude, $s > 0$ is the guidance strength, and $\epsilon(t)$ represents the accumulated errors due to non-linearity and trajectory crossovers.

\begin{proof}

Consider the sampling process described by the ODE:

\begin{equation}
    \frac{dx(t)}{dt} = f(x(t),t),
\end{equation}

where $f(x(t),t)$ is a nonlinear function often parameterized via neural network $\theta$. 
To guide the sampling process toward minimizing a loss function $\mathcal{L}(\hat{x}_0,x_0^{ref})$, we can adjust the dynamics by adding the gradient $\nabla_{x_t}$ to the vector field (see Eq.~\ref{eq:baseline_control}) as: 

\begin{equation} 
    \frac{dx(t)}{dt} = f(x(t), t) - s \cdot \nabla_{x_t} \mathcal{L}(\hat{x}_0, x_0^{\text{ref}}), 
\end{equation}

where $s$ is the guidance strength. 
Let $e(t) = \hat{x}_0 - x_0^{ref}$ be the error between the estimated and target samples. 
Since $\hat{x}_0(t) = x(t) + \int_t^0 f(x(\tau), \tau) d\tau$, differentiating $e(t)$ with respect to $t$ yields:

\begin{align} 
\frac{d e(t)}{dt} &= \frac{d \hat{x}_0(t)}{dt} \\ &= \frac{d x(t)}{dt} - f(x(t), t) \\ &= - s \cdot \nabla_{x_t} \mathcal{L}(\hat{x}_0, x_0^{\text{ref}}). 
\end{align}

However, this requires the compute-intensive backpropagation through ODESolver. 
Therefore, it is important to find an approximation of $\nabla_{x_t}$.
And the most convenient approximation is: $\nabla_{x_t} \approx \nabla_{\hat{x}_0}$.
However, this derivation assumes that the integral $\int_t^0 f(x(\tau), \tau) d\tau$ is well-behaved and that $\hat{x}_0(t)$ depends smoothly on $x(t)$. 
In the presence of nonlinearity and trajectory crossovers, small changes in $x(t)$ can lead to disproportionately large changes in $\hat{x}_0(t)$, due to the sensitivity of the integral to the path taken.
Moreover, potential crossovers in the trajectory mean that the mapping from $x(t)$ to $\hat{x}_0(t)$ is not injective; different trajectories $x(t)$ may lead to the same $\hat{x}_0(t)$ or vice versa. 
This non-unique mapping complicates the error dynamics because $\nabla_{\hat{x}_0}\mathcal{L}$ may not provide a consistent or effective direction for updating $x(t)$.
Including the effects of nonlinearity and trajectory crossovers, the error dynamics become:

\begin{equation} 
    \frac{d e(t)}{dt} = - s \cdot \nabla_{\hat{x}_0} \mathcal{L}(\hat{x}_0, x_0^{\text{ref}}) + \epsilon(t), 
\end{equation}

where $\epsilon(t)$ represents the errors introduced by the nonlinearity in $f(x(t),t)$ and the sensitivity of $\hat{x}_0$ to $x(t)$ due to trajectory crossovers. 
In other words, the approximation error $\epsilon(t)$ can be represented as:

\begin{equation}
    \epsilon(t) = s \cdot \left(\nabla_{x_t} \mathcal{L}(\hat{x}_0, x_0^{ref})- \nabla_{\hat{x}_t} \mathcal{L}(\hat{x}_0, x_0^{ref}) \right).
\end{equation}

Assuming a quadratic loss function $\mathcal{L} = ||\hat{x}_0 - x_0^{ref}||_2^2$, we have $\nabla_{\hat{x}_0}\mathcal{L} = 2e(t)$, leading to:

\begin{equation} 
    \frac{d e(t)}{dt} = - 2 s e(t) + \epsilon(t).     
\end{equation}

To understand the convergence of the error, we analyze the evolution of the error magnitude $E(t) = e(t)^T e(t)$. 
Differentiating $E(t)$ with respect to time $t$, we get:

\begin{align} 
\frac{dE(t)}{dt} &= \frac{d}{dt} \left( e(t)^\top e(t) \right) \\
&= 2 e(t)^\top \frac{d e(t)}{dt} \\
&= 2 e(t)^\top \left( - 2 s e(t) + \epsilon(t) \right) \\
&= - 4 s e(t)^\top e(t) + 2 e(t)^\top \epsilon(t) \\
&= - 4 s E(t) + 2 e(t)^\top \epsilon(t). 
\end{align}

This completes the proof.
\end{proof}

Notably, we derive this behavior of the ODE processes under the assumption that the error rate cannot be calculated accurately. 
This can either come from the incorrect estimation of $\hat{x}_0$ or the nonlinearity of ODESolver itself. 
In the next section, we further concretize this with respect to the RFMs.

\section{Proof for Theorem}
\label{sec:appendix_theorem}

\paragraph{Lemma 4.2} (Gradient Relationship)\textbf{.}
    Let $u_\theta : \mathcal{R}^d \times [0,T] \rightarrow \mathcal{R}^d$ be the velocity function with the parameter $\theta$.
    Then the gradient of the cost function $\mathcal{L}$ at any timestep $t$ can be approximated as:

    \begin{equation}
        \nabla_{x_t} \mathcal{L} = (I + t \cdot J_{u_\theta})^T \nabla_{\hat{x}_0} \mathcal{L}.
        \label{eq:lemma_appendix}
    \end{equation}

\begin{proof}
    Leveraging the straight-line trajectories characteristic of rectified flow models, the data sample at $t=0$ can be estimated directly from an intermediate state $x_t$:

    \begin{equation}
        \hat{x}_0 = x_t + t \cdot u_\theta(x_t, t).
    \end{equation}
    
    By differentiating the $\hat{x}_0$ with respect to $x_t$, we get:

    \begin{align}
        \frac{d\hat{x}_0}{dx_t} &= I + t \cdot \frac{du_\theta(x_t, t)}{dx_t}\\
        &=I + t \cdot J_{u_\theta}(x_t, t).
    \end{align}

    Using the chain rule for gradients:

    \begin{equation}
        \nabla_{x_t} \mathcal{L} = \left( \frac{d\hat{x}_0}{dx_t} \right)^T \nabla_{\hat{x}_0} \mathcal{L}.
    \end{equation}

    Substituting the expression for $\frac{d\hat{x}_0}{dx_t}$, we obtain:

    \begin{equation}
        \nabla_{x_t} \mathcal{L} = (I + t \cdot J_{u_\theta}(x_t, t))^T \nabla_{\hat{x}_0}\mathcal{L}.
    \end{equation}

    According to Assumption 3, due to the constancy of Jacobian, $J_{u_\theta}$, for rectified flow models, we can treat it as constant for any time $t$.
    Hence, we get our desired approximation:

    \begin{equation}
        \nabla_{x_t} \mathcal{L} = (I + t \cdot J_{u_\theta}(x_t, t))^T \nabla_{\hat{x}_0}\mathcal{L}.
    \end{equation}

    This completes the proof.

\end{proof}

Hence, as either $t \rightarrow 0$ or $J_{u_\theta}$ (Assumption 2), both gradients become approximately similar and $\epsilon(t) \rightarrow 0$. 
This guarantees the convergence of the error dynamics as time passes.
We further show this behavior of RFMs empirically in Section~\ref{sec:appendix_findings} and show that this remains true for even large-scale latent models.

\paragraph{Theorem 4.3} (Update Rule for Steering the RFMs)\textbf{.}
Let $u_\theta: \mathbb{R}^d \times [0, T] \rightarrow \mathbb{R}^d$ be a velocity field with constant Jacobian $J_{u_\theta}$. Define the estimated initial state $\hat{x}_0$ from an intermediate state $x_t$ by
\begin{equation*}
    \hat{x}_0 = x_t + t \cdot u_\theta(x_t, t).
\end{equation*}
Consider the quadratic loss function $\mathcal{L} = \|\hat{x}_0 - x_0^{\text{ref}}\|^2$, where $x_0^{\text{ref}}$ is a reference sample. Then, the update rule for controlled generation is given by
\begin{equation*}
    x_{t - \Delta t} = x_t + \Delta t u_\theta(x_t, t) - s' \nabla_{\hat{x}_0} \mathcal{L},
\end{equation*}
where:
\begin{itemize}
    \item $\nabla_{\hat{x}_0} \mathcal{L} = 2(\hat{x}_0 - x_0^{\text{ref}})$,
    \item $s' \approx \left( I + \Delta t \cdot J_{u_\theta} \right) \left( I + t \cdot J_{u_\theta} \right)^\top$,
    \item $I$ is the identity matrix.
\end{itemize}

\begin{proof}
    By lemma~\ref{lemma:grad_approx} and Assumption 2, we can further approximate the Eq.~\ref{eq:lemma_appendix}:

    \begin{equation}
        \nabla_{x_t} \mathcal{L} = (I + t \cdot J_{u_\theta})^T \nabla_{\hat{x}_0} \mathcal{L} \approx K^T \nabla_{\hat{x}_0} \mathcal{L},
    \end{equation}

    where $K$ is the constant matrics as $\Delta t \rightarrow 0$ and $t \rightarrow 0$. Under this formulation, we can perform controlled image generation in three steps:

    \begin{equation}
    \begin{aligned}
        &\text{Step 1:} \quad & \hat{x}_0 &= x_t + t \cdot u_\theta(x_t, t) \\
        &\text{Step 2:} \quad & \hat{x}_t &= x_t - K^T\nabla_{\hat{x}_0}\mathcal{L} \\
        &\text{Step 3:} \quad & x_{t-\Delta t} &=\hat{x}_t + \Delta t \cdot u_\theta(\hat{x}_t, t).
    \end{aligned}
    \end{equation}

    However, this will require additional forward passes.
    But according to Assumption 2 if $\Delta t$ is sufficiently small, then by Taylor series approximation, we get:

    \begin{align}
    x_{t-\Delta t} &= x_t - K^T \nabla_{\hat{x}_0} \mathcal{L} 
                     + \Delta t \cdot u_\theta \left( x_t - K^T \nabla_{\hat{x}_0} \mathcal{L}, t \right) \\
                   &= x_t - K^T \nabla_{\hat{x}_0} \mathcal{L} \notag \\
                   &\quad + \Delta t \left[ u_\theta(x_t, t) - J_{u_\theta} \cdot K^T \cdot \nabla_{\hat{x}_0} \mathcal{L} \right]
    \end{align}

    Now, as $J_{u_\theta}$ is constant \textit{w.r.t.} $\Delta t$.
    Hence, we get:

    \begin{align}
    x_{t-\Delta t} &= x_t - (I + \Delta t \cdot J_{u_\theta}) K^T \nabla_{\hat{x}_0} \mathcal{L} 
                     + \Delta t \cdot u_\theta(x_t, t)\\
                 &= x_t + \Delta t \cdot u_\theta(x_t, t) - s' \nabla_{\hat{x}_0}\mathcal{L},
    \end{align}
    where $s' = (I + \Delta t \cdot J_{u_\theta}) K^T$ is constant and it can predetermined.

    Hence, this concludes the proof that for appropriate guidance scale $s'$, we can perform the controlled generation as derived above.
\end{proof}

\section{Numerical Accuracy for Model Steering}
\label{sec:appendix_numerical_accuracy}

In our controlled generation framework, we aim to steer the generation process towards a reference sample $x_0^{ref}$ by solving the modified ODE:
\begin{equation}
    \frac{d x(t)}{dt} = f(x(t), t) = u_\theta(x(t), t) - s' \nabla_{\hat{x}_0} \mathcal{L}.
    \label{eq:controlled_ivp}
\end{equation}

The accuracy of this numerical integration is crucial, as errors can accumulate over time, leading to deviations from the desired trajectory. The smoothness of the modified velocity field $f(x(t), t)$ significantly impacts this accuracy. 
Specifically, a smaller magnitude of $\left| \frac{d}{dt} f(x(t), t) \right|$ reduces local truncation errors.
The following Proposition formalizes this relationship, stating that the numerical accuracy improves as $\left| \frac{d}{dt} f(x(t), t) \right|$ decreases.

\begin{proposition} (Informal)\textbf{.}
Given the prior notations, Assumptions, and Theorem, 
for any $p$-th order numerical method solving Eq.~\eqref{eq:controlled_ivp}, the accuracy of the numerical solution increases as the magnitude of $\left| \frac{d}{dt} f(x(t), t) \right|$ decreases.
\end{proposition} 

\begin{proof}

To analyze the local truncation error, consider the Taylor series expansion of the exact solution around time $t$ when integrating backward in time from $t$ to $t - \Delta t$:

\begin{align*}
\label{eq:taylor_exact}
x(t - \Delta t) = &x(t) - \Delta t\, f(x(t), t) + \frac{(\Delta t)^2}{2} \frac{d}{dt} f(x(t), t) \\&- \frac{(\Delta t)^3}{6} \frac{d^2}{dt^2} f(x(t), t) + O\left((\Delta t)^4\right).
\end{align*}

The numerical method updates the solution using:

\begin{equation}
\label{eq:numerical_update}
x_{t - \Delta t} = x_t + \Delta t\, \phi(x_t, t),
\end{equation}

where $\phi(x_t, t)$ is the increment function.
The local truncation error $\tau$ is the difference between the exact solution and the numerical approximation:

\begin{align*}
\tau &= x(t - \Delta t) - x_{t - \Delta t} \\
&= \left[ x(t) - \Delta t\, f(x(t), t) + \frac{(\Delta t)^2}{2} \frac{d}{dt} f(x(t), t) \right. \\
&\quad \left. - \frac{(\Delta t)^3}{6} \frac{d^2}{dt^2} f(x(t), t) + O\left((\Delta t)^4\right) \right] \\
&\quad - \left[ x_t + \Delta t\, \phi(x_t, t) \right].
\end{align*}

The first p-order terms cancel out, and we have:

\begin{equation}
||\tau|| \le \left\|\frac{(\Delta t)^{p+1}}{(p+2)!}  \frac{d^{p+1}}{dt^{p+1}} f(x(t),t)\right\|
\end{equation}

According to the Mean Value Theorem, we have
\begin{equation}
||\tau||\le C (\Delta t)^{p+1}\max_{t \in [t_n, t_{n+1}]} \left\|\frac{d}{dt} f(x(t),t) \right\|
\end{equation}

where $C$ is a constant depending on the method.
The global error $e(t) = x(t) - x_t$ accumulates these local errors over the integration interval. Under standard assumptions (e.g., Lipschitz continuity of $f$), the global error is bounded by:

\begin{equation}
\label{eq:global_error_bound}
\|e(t)\| \le K (\Delta t)^{p} \left( e^{L (T - t)} - 1 \right) \max_{t \in [0, T]} \left\| \frac{d}{dt} f(x(t),t) \right\|,
\end{equation}

where $K$ is a constant depending on the Lipschitz constant $L$ of $f$ and the total integration time $T$.
\end{proof}

As the magnitude of $\left\| \frac{d}{dt} f(x_t, t) \right\|$ decreases, both the local truncation error and the global error decrease, enhancing the accuracy of the numerical solution. In the context of controlled generation, ensuring that $f(x_t, t)$ changes smoothly over time leads to more accurate integration and better alignment with the reference point $x_0^{\text{ref}}$.
This insight and prior assumptions require that the guidance scale $s'$ and $\delta t$ be sufficiently smaller, where higher NFEs lead to the lower $\Delta t$.
Hence, we increase the NFEs significantly to stabilize the steering (see Section~\ref{sec:appendix_hyperparams}).
By carefully selecting $s'$, we ensure that the additional term $s' \cdot \nabla_{\hat{x}_0} \mathcal{L}$ does not introduce excessive variability into $f(x(t),t)$, maintaining the smoothness necessary for accurate numerical integration.

\section{Extended Related Works}
\label{sec:appendix_related}

\paragraph{Generative Models.}
Recent advances in generative models, especially diffusion models like Latent Diffusion Model (LDM)~\cite{rombach2022high}, GLIDE~\cite{nichol2021glide}, and DALL-E2~\cite{ramesh2022hierarchical}, have significantly improved photorealism compared to GAN-based methods such as StackGAN~\cite{zhang2017stackgan} and BigGAN~\cite{brock2018large}. Pretrained diffusion models have been successfully applied to diverse tasks, including image editing~\cite{hertz2023prompttoprompt}, personalization~\cite{patel2024lambda}, and style transfer~\cite{wang2024instantstyle}, but their inference flexibility remains limited, and they demand substantial resources~\cite{gu2024matryoshka, patel2024eclipse}. Distillation-based strategies like Latent Consistency Models~\cite{luo2023latent} and Distribution Matching Distillation~\cite{yin2024one} address some limitations but lack control and broader applicability. Rectified Flow Models (RFMs)~\cite{liu2023flow, lipman2023flow}, exemplified by Flux\footnote{\url{https://huggingface.co/black-forest-labs/FLUX.1-dev}}, SD3~\cite{esser2024scaling}, and InstaFlow~\cite{liu2023instaflow}, show promise but face challenges in downstream tasks due to inversion inaccuracies and other limitations. This work addresses these gaps, extending RFMs to downstream tasks in a training-, gradient-, and inversion-free manner.

\paragraph{Conditional Sampling.}
Song \textit{et al.} introduced noise-aware classifiers for controlling sampling in diffusion models~\cite{dhariwal2021diffusion}, but these require task-specific training. Classifier-free guidance (CFG)~\cite{ho2022classifier} avoids this but necessitates an additional pretraining stage. FreeDoM~\cite{yu2023freedom} and MPGD~\cite{hemanifold} improve sampling control but remain computationally intensive. Initial extensions of conditional sampling to flow models face similar challenges, such as compute-heavy gradient backpropagation and limited applicability to latent space models. Our method, \flowchef, eliminates these issues, seamlessly enabling gradient- and inversion-free conditional sampling in latent-space models.

\paragraph{Inverse Problems.}
Inverse problems, dominated by diffusion-based methods~\cite{daras2024survey}, include pixel-space solutions such as DPS~\cite{chung2022diffusion}, $\Pi$-GDM~\cite{peng2024improving}, and BlindDPS~\cite{chung2023parallel}. PSLD~\cite{rout2024solving} extends support to latent-space models, while manifold-based methods~\cite{songsolving, hemanifold} further enhance performance. Flow-based approaches like OT-ODE~\cite{pokle2024trainingfree} and D-Flow~\cite{ben-hamu2024dflow} improve speed and quality but remain resource-intensive. Recent advancements like PnP-Flow~\cite{martin2024pnp} achieve training- and gradient-free solutions for pixel-space models but face issues like smoothness artifacts. Existing solutions are resource-intensive and unsuitable for large-scale latent models. \flowchef~leverages vector field properties of RFMs to enhance performance, generalization, and scalability for state-of-the-art models like Flux.

\paragraph{Image Editing.}
Image editing typically involves guiding a model to combine a reference image with an editing instruction, often through inversion~\cite{mokady2023null, hertz2023prompttoprompt, huberman2024edit, ju2023direct}. Inversion-free methods like DiffEdit~\cite{couairon2023diffedit}, InfEdit~\cite{xu2023inversion}, and TurboEdit~\cite{wu2024turboedit} are rare, and none apply to flow models. Most state-of-the-art methods rely on cross-attention mechanisms~\cite{brack2024ledits++, mokady2023null}, which we do not prioritize. Our approach, \flowchef, introduces the first inversion-free image editing method for RFMs, achieving competitive results with state-of-the-art methods.

\section{Empirical Findings}
\label{sec:appendix_findings}

\begin{figure*}[ht!]
    \centering
    \subfloat[Gradient Similarity in InstaFlow \textit{vs.} Stable Diffusion v1.5.]{%
        \includegraphics[width=0.23\textwidth]{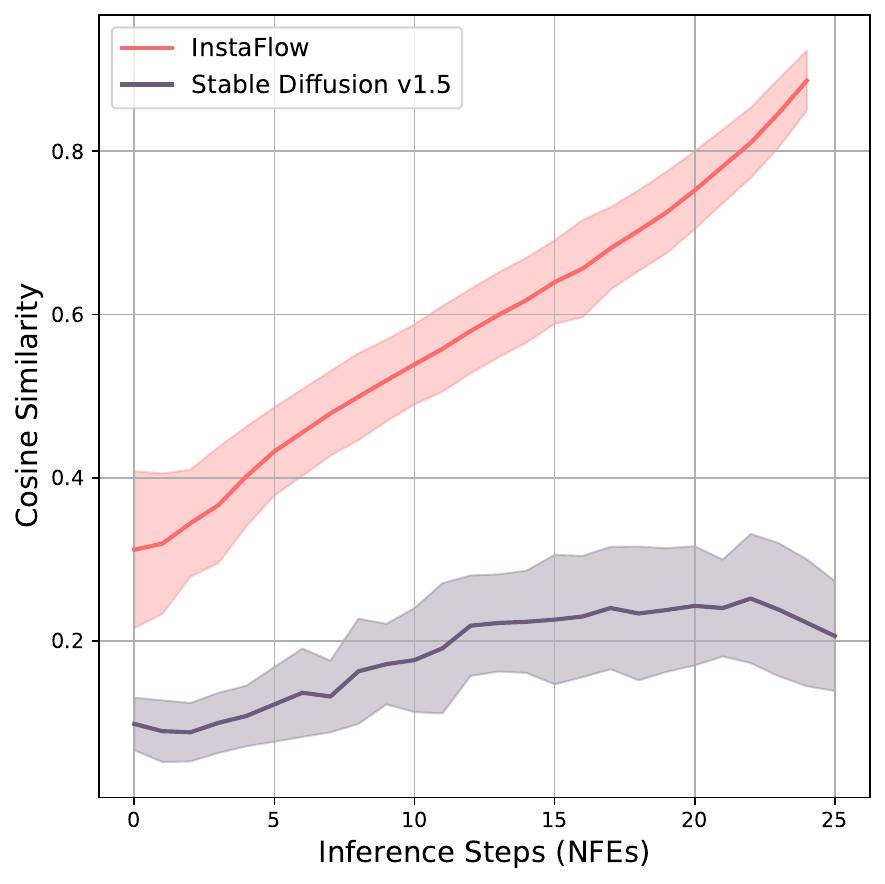} 
        \label{fig:empirical_fig1}
    }
    \hfill
    \subfloat[Gradient Similarity in Rectified Flow ++ model.]{%
        \includegraphics[width=0.23\textwidth]{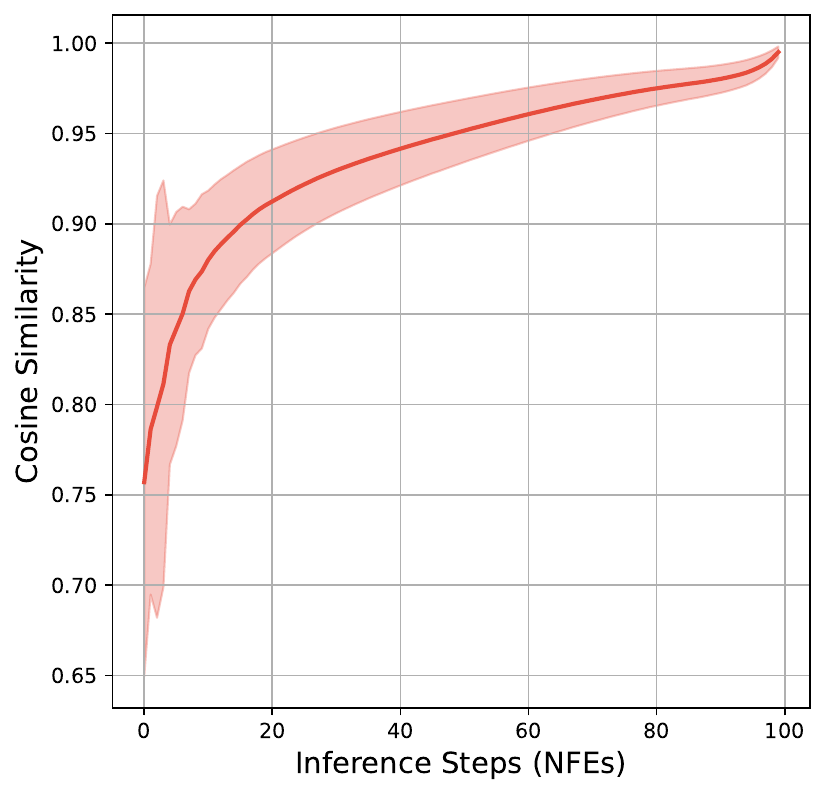} 
        \label{fig:empirical_fig2}
    }
    \hfill
    \subfloat[Gradient Similarity in Rectified Flow ++ during model steering.]{%
        \includegraphics[width=0.23\textwidth]{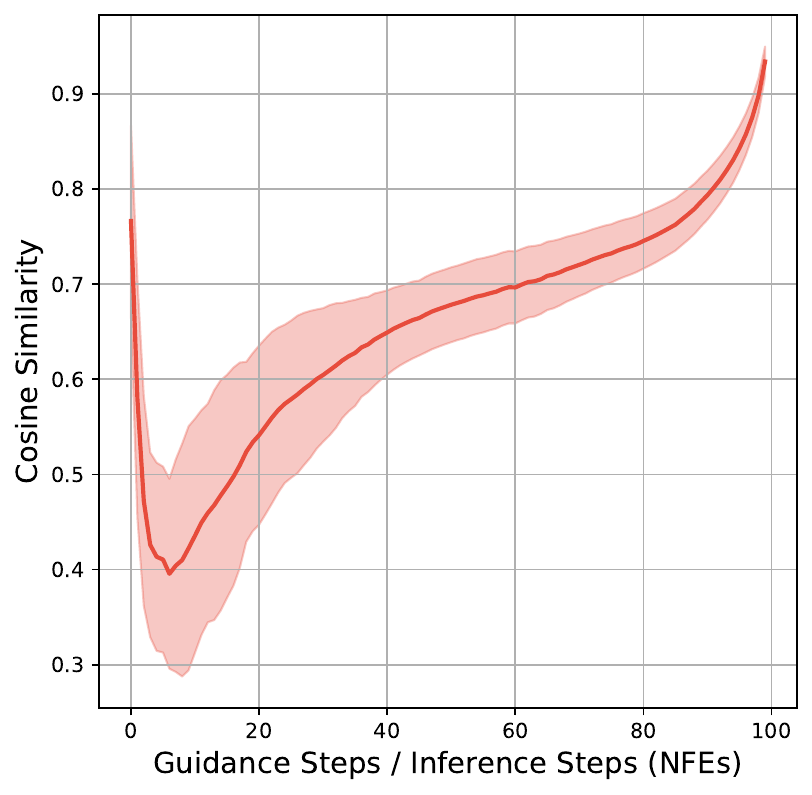} 
        \label{fig:empirical_fig3}
    }
    \hfill
    \subfloat[Convergence in Rectified Flow ++ during model steering.]{%
        \includegraphics[width=0.23\textwidth]{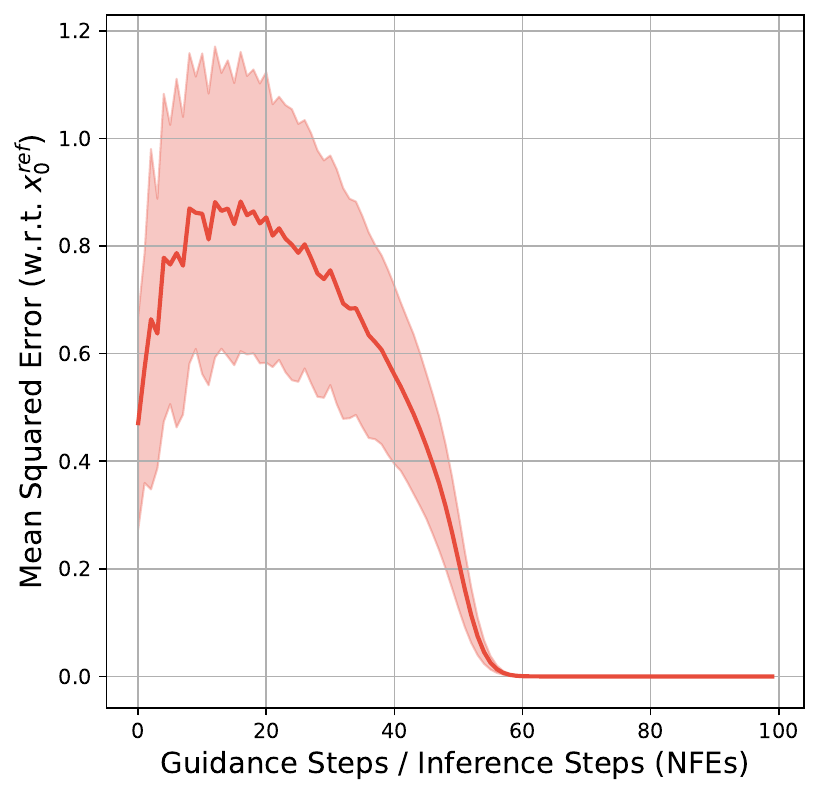} 
        \label{fig:empirical_fig4}
    }
    
    \caption{Empirical analysis of gradient similarity (a, b, and c) and convergence rate. (a) and (b) analyzes the gradients without model steering. (c) contains the gradient similarity during the active model steering. And (d) shows the trajectory similarity at each timestep $t$ \textit{w.r.t.} the inversion based trajectory.}
    \label{fig:empirical_analysis}
\end{figure*}

In Section~\ref{sec:method}, we provided theoretical insights into \flowchef~along with an intuitive algorithm. To complement the theory, we conducted an empirical analysis on large-scale RFMs to validate the Assumptions, Propositions, Lemmas, and Theorems presented. The results are summarized in Figure~\ref{fig:empirical_analysis}.

In Figure~\ref{fig:empirical_fig1}, we compare the gradient cosine similarity with and without backpropagation through the ODESolver for InstaFlow and Stable Diffusion v1.5. For all denoising steps, the gradients of SDv1.5 behave nearly randomly, indicating that the stochasticity of the base model significantly impacts gradients, even when using the ODE sampling process during inference. In contrast, for InstaFlow, as denoising progresses ($t \rightarrow 0$), gradient alignment improves, supporting our derivation in Lemma~\ref{lemma:grad_approx}, which states that as $t \rightarrow 0$, we achieve $\nabla_{x_t} \approx \nabla_{\hat{x}_0}$.

Further analysis was performed on the Rectified Flow++ model, which is designed for straight trajectories with zero crossovers. As shown in Figure~\ref{fig:empirical_fig2}, well-trained models exhibit high gradient similarity even at the initial stages of denoising. However, as illustrated in Figure~\ref{fig:empirical_fig3}, during active steering, the gradient direction initially diverges before improving. This behavior is also reflected in the convergence plot in Figure~\ref{fig:empirical_fig4}. 

We hypothesize that this phenomenon arises due to the proximity to the Gaussian noise space ($p_1 \sim N(0, I)$), where model steering is more error-prone since minor adjustments can disproportionately affect future trajectories. As denoising progresses and the distribution moves further from the noise ($p_1$), these errors diminish, and convergence is achieved. These observations align well with our theoretical predictions, further reinforcing the validity of \flowchef.

\section{Algorithms}
\label{sec:appendix_algos}

\begin{algorithm}[!t]
\DontPrintSemicolon
\textbf{Input:} Pretrained Rectified-flow model $u_\theta$, input noise sample $x_T\sim N(0, I)$, target data sample $x_0^{ref}$, and $\mathcal{L}$ cost function. \;

\For {$t \in \{T...0\}$}
{
 \colorbox{ForestGreen!30}{$v \leftarrow u_\theta(x_t, t)$ \;}
 $dt \leftarrow 1/T$ \;
 $x_t \leftarrow x_t.require\_grad\_(True)$ \;
 
  \For {N \textnormal{steps}}
    {  
    \colorbox{red!30}{$v \leftarrow u_\theta(x_t, t)$ \;}\\
    $\hat{x}_0 \leftarrow x_t + t\cdot v$
    
    $loss \leftarrow \mathcal{L}(\hat{x}_0, 
    x_0^{ref})$
    
    \colorbox{red!30}{$\nabla_{x_t} \leftarrow$ grad(loss, $x_t$) \quad // Compute heavy BP}
    
    \colorbox{ForestGreen!30}{$x_t \leftarrow$ Optimize($x_t$, loss) \quad \quad \quad // Lemma~\ref{lemma:grad_approx}}
    }

    \colorbox{red!30}{$v \leftarrow u_\theta(x_t, t)$ \;}\\
    $x_{t-1} \leftarrow x_t + dt\cdot v$ \quad~~~ \quad \quad \quad \quad // Theorem~\ref{theorem:method}

}

\textbf{RETURN} $x_0$
\caption{\flowchef~\textit{vs.} Baseline FreeDoM.}
\label{algo:algo2}
\end{algorithm}

\begin{algorithm}[!t]
\DontPrintSemicolon
\textbf{Input:} Pretrained Rectified-flow model $u_\theta$, input noise sample $x_T\sim N(0, I)$, target data sample $x_0^{ref}$, $c^{edit}$ is edit prompt, $c^{base}$ is base prompt, $M$ is user-provided input mask, and $\mathcal{L}$ cost function. \;

\For {$t \in \{T...0\}$}
{
 $dt \leftarrow 1/T$ \;
 $c \leftarrow [c^{edit}, c^{base}]$ \;
 $v \leftarrow u_\theta(x_t, t, c)$ \;
$v_{edit}, v_{base} = v.chunk(2)$ \;
$v = v_{edit} + \neg mask \cdot (v_{edit} - v_{base}) \cdot s$ \;
$M_{edit} \leftarrow M$\;
 $x_t \leftarrow x_t.require\_grad\_(true)$ \;

\If{$t < min_T$}
{
  \For {N \textnormal{steps}}
    {  
    $\hat{x}_0 \leftarrow x_t + t\cdot v$

    \If{$t < max\_full\_steps_T$}
    {
        $M_{edit} \leftarrow I$
    }
    
    $loss \leftarrow \mathcal{L}(\hat{x}_0, 
    x_0^{ref})\cdot M_{edit}$ \;
    
    $x_t \leftarrow$ Optimize($x_t$, loss)\quad // Lemma~\ref{lemma:grad_approx}
    } 

}

    $x_{t-1} \leftarrow x_t + dt\cdot v$ \quad \quad \quad \quad \quad // Theorem~\ref{theorem:method}

}

\textbf{RETURN} $x_0$
\caption{: \flowchef~optimized for a wide range of image editing tasks.}
\label{algo:algo3}
\end{algorithm}

This section provides an overview of the algorithms underpinning \flowchef~for image editing and its comparison to baseline methods for a comprehensive understanding.

\paragraph{Image Editing.}
As described in Section~\ref{sec:flowchef}, \flowchef~can be easily extended to image editing. Revisiting the core concept, \flowchef~modifies random trajectories to align with a target sample. Image editing involves balancing similarity with the target sample while introducing deviations to achieve desired edits.

Figure~\ref{fig:algo_hyperparameter} and Section~\ref{sec:appendix_hyperparams} illustrate how \flowchef~progressively transfers characteristics from high-level structure to finer details like color composition. However, editing requirements vary by task. For example, adding an object benefits from trajectory adjustments earlier in the denoising process, while color changes require gradual learning at later stages. We can optimize parameters for diverse tasks using the generalized \flowchef, as detailed in Algorithm~\ref{algo:algo1}.

To simplify the process, we extend \flowchef~to support off-the-shelf editing tasks, such as those in the PIE-Benchmark, as detailed in Algorithm~\ref{algo:algo2}. Assume a non-edit region mask, $M_{edit}$, derived from cross-attention or human annotation. To steer the trajectory towards the desired edits, we modify the velocity ($v$) using a classifier-free guidance strategy:

\begin{equation}
v = v_{edit} + \neg mask \cdot (v_{edit} - v_{base}) \cdot s,
\end{equation}

where $v_{edit}$ corresponds to the edit prompt and $v_{base}$ to the base (negative) prompt. This adjustment ensures the trajectory reflects the desired edits.

To maintain alignment of non-edited regions with the target sample, we modify the cost function as follows:

\begin{equation}
    \mathcal{L}(\hat{x}_0, x_0^{ref}) = ||(\hat{x}_0 - x_0^{ref})\cdot M_{edit}||_2^2.
\end{equation}

Preserving the original image structure is crucial for edits such as color or material changes. To achieve this, we introduce the parameter $max\_full\_steps\_T$, which determines the number of steps that apply full \flowchef~guidance with an identity mask. This ensures structural preservation while facilitating edits. Section~\ref{sec:appendix_hyperparams} details a comprehensive reference for hyperparameters.

\paragraph{\flowchef~vs. Baseline FreeDoM.}
Algorithm~\ref{algo:algo2} compares \flowchef~to the baseline FreeDoM, a diffusion model method that modifies the score function using a classifier guidance-like approach. FreeDoM requires estimating velocity and calculating gradients ($\nabla_{x_t}$) through backpropagation via the ODESolver $u_\theta$, as marked in red. In contrast, as highlighted in green, \flowchef~eliminates the need for backpropagation while still achieving convergence. This simplification makes \flowchef~a more efficient and practical solution without sacrificing performance.

\section{Experimental Setup}
\label{sec:appendix_setup}

\begin{table}[!t]
    \centering
    \captionsetup{font=small}
    \scriptsize

    \resizebox{\linewidth}{!}{
    \begin{tabular}{l ccccc}
        \toprule
        \textbf{Hyperparameter} &\textbf{OT-ODE} &\textbf{D-Flow} &\textbf{PnP-Flow} &\textbf{FlowChef} \\\midrule
        \textbf{Iterations / NFEs} &200 &20 &50 &200 \\
        \textbf{Optimization per iteration} &1 &- &- &1 \\
        \textbf{Optimization per denoising} &- &50 &- &- \\
        \textbf{Avg. sampling steps} &- &- &5 &- \\
        \textbf{Guidance scale} &1 &1 &1 &500 \\
        \textbf{Cost function} &L1 &L1**2  &L1 &MSE \\
        \textbf{initial time (1 means noise)} &0.8 &- &- &- \\
        \textbf{blending strength} &- &0.05 &- &- \\
        \textbf{inversion} &$\times$ &$\checkmark$ &$\times$ &$\times$ \\
        \textbf{learning rate} &1 &1 &1 &1 \\
        \bottomrule
    \end{tabular}
    }
    \caption{Hyperparameters for solving inverse problems using pixel-space models.}
    \label{tab:hyperpara_pixels}
\end{table}

\begin{table}[!t]
    \centering
    \captionsetup{font=small}
    \scriptsize

    \resizebox{\linewidth}{!}{
    \begin{tabular}{l ccc}
        \toprule
        \textbf{Hyperparameter} &\textbf{D-Flow} &\textbf{RectifID} &\textbf{FlowChef} \\\midrule
        \textbf{Iterations / NFEs} &10 &4 &100 \\
        \textbf{Optimization per iteration} &- &- &1 \\
        \textbf{Optimization per denoising} &20 &400 &- \\
        \textbf{Blending strength} &0.1 &- &- \\
        \textbf{Guidance scale} &0.5 &0.5 &0.5 \\
        \textbf{Cost function} &MSE &MSE &MSE \\
        \textbf{Learning rate} &0.5 &1 &0.02 \\
        \textbf{Optimizer} &Adam &SGD &Adam \\
        \textbf{loss multiplier (latent/pixel)} &0.000001 &0.0001 / 100000 &0.001/1000 \\
        \textbf{inversion} &$\checkmark$ &$\times$ &$\times$ \\
        \bottomrule
    \end{tabular}
    }
    \caption{Hyperparameters for solving inverse problems using latent-space models (InstaFlow).}
    \label{tab:hyperpara_latents}
\end{table}

This section outlines the hyperparameters used for \flowchef~and baseline methods in solving inverse problems.

\paragraph{Pixel-Space Models.}
All evaluations were conducted using the Rectified Flow++ checkpoint. Since public implementations of OT-ODE and D-Flow are unavailable, we implemented these methods manually based on the provided pseudocode and performed hyperparameter tuning to ensure optimal performance. 
Notably, DPS and FreeDoM hyperparameters are the same as the \flowchef.
Table~\ref{tab:hyperpara_pixels} provides a detailed overview of the hyperparameters used for each baseline.

\begin{table*}[!t]
    \centering
    \captionsetup{font=small}
    \scriptsize

    \resizebox{\textwidth}{!}{
    \begin{tabular}{ll ccccccccc}
        \toprule
        \textbf{Model} &\textbf{Hyperparameters} &\textbf{Chage Object} &\textbf{Add Object} &\textbf{Remove Object} &\textbf{Change Attrbiute} &\textbf{Chage Pose} &\textbf{Change Color} &\textbf{Change Material} &\textbf{Change Background} &\textbf{Change Style} \\\midrule
        \multirow{6}{*}{\textbf{FlowChef (InstaFlow)}} &\textbf{Learning rate} &0.5 &0.5 &0.5 &0.5 &0.5 &0.5 &0.5 &1.0 &0.5 \\
        &\textbf{Max setps} &50 &50 &50 &50 &20 &30 &50 &50 &30 \\
        &\textbf{Optimization steps} &1 &1 &3 &2 &2 &2 &2 &4 &1 \\
        &\textbf{Inference steps} &50 &50 &50 &50 &50 &50 &50 &50 &50 \\
        &\textbf{Full source steps} &30 &30 &0 &10 &10 &20 &20 &0 &30 \\
        &\textbf{Edit guidance scale} &2.0 &2.0 &2.0 &4.5 &8.0 &8.0 &4.0 &3.0 &6.0 \\
        \midrule
        
        \multirow{6}{*}{\textbf{FlowChef (Flux)}}
        &\textbf{Learning rate} &0.4 &0.5 &0.5 &0.5 &0.5 &0.4 &0.5 &0.5 &0.4 \\
        &\textbf{Optimization steps} &1 &1 &1 &1 &1 &1 &1 &1 &1 \\
        &\textbf{Inference steps} &30 &30 &30 &30 &30 &30 &30 &30 &30 \\
        &\textbf{Full source steps} &5 &5 &0 &2 &5 &3 &5 &0 &5 \\
        &\textbf{Edit guidance scale} &4.5 &4.5 &4.5 &4.5 &7.5 &10.0 &4.5 &0.0 &10.0 \\
        \bottomrule
    \end{tabular}
    }
    \caption{Hyperparameter examples for which various editing tasks can be performed (following Algorithm 2). Notably, the \flowchef~(Flux) variant can be further optimized for task-specific settings that will follow Algorithm 1 with a careful selection of hyperparameters.}
    \label{tab:hyperpara_edit}
\end{table*}

\paragraph{Latent-Space Models.}
For latent-space models, we extended D-Flow to the InstaFlow pretrained model, repurposed RectifID for inverse problems, and fine-tuned the hyperparameters for optimal results. The best-performing hyperparameters for each baseline are listed in Table~\ref{tab:hyperpara_latents}. 
We utilized their baseline implementations for diffusion model-based approaches such as Resample and PSLD-LDM, modifying only the number of inference steps. Specifically, we used 100 NFEs for Resample and 100/500 NFEs for PSLD.



\section{RF-Inversion \textit{vs.} \flowchef}
\label{sec:appendix_rfinversion}
\begin{table}[!t]
    \centering
    \captionsetup{font=small}
    \scriptsize

    \resizebox{\linewidth}{!}{
    \begin{tabular}{l|ccc}
        \toprule
        \textbf{Methods} & \textbf{CLIP-I} ($\uparrow$) & \textbf{CLIP-T} ($\uparrow$) & \textbf{Time} ($\downarrow$)\\
        \midrule
        RF-Inversion                 & \cellcolor{ForestGreen!30}\textbf{0.8573}   & 0.2790  & $\sim$31 sec \\
        \cellcolor{Apricot!50} \textbf{\flowchef~(ours)} & 0.8269 & \cellcolor{ForestGreen!30}\textbf{0.2828} & \cellcolor{ForestGreen!30}\textbf{$\sim$15 sec} \\
        \bottomrule
    \end{tabular}
    }
    \caption{Comparison of \flowchef~with concurrent work RF-Inversion on top of Flux for editing task ``wearing glasses''.}
    \label{tab:wearing_glasses}
    
\end{table} 

In this section, we briefly compare \flowchef~with the concurrent work, RF-Inversion, which introduces an inversion strategy for rectified flow models using a linear quadratic regulator perspective from optimal transport, particularly for image editing tasks.
RF-Inversion relies on image inversion, significantly increasing compute time—nearly doubling it compared to \flowchef. To evaluate, we conducted a ``wearing glasses'' editing task using 256 randomly selected SFHQ face images on the Flux.1[dev] model. As shown in Table~\ref{tab:wearing_glasses}, \flowchef~achieves competitive performance in half the time.
At a high level, RF-Inversion can be viewed as a special case of \flowchef, where the starting point is an inverted image rather than random noise. We applied a similar editing strategy to both methods for a fair comparison, as outlined in Algorithm~\ref{algo:algo1}, using a learning rate of 0.07, 20 optimization steps, and 30 total inference steps. On an A100 GPU, this configuration required approximately 15 seconds per inference. This comparison highlights the efficiency and versatility of \flowchef~in handling image editing tasks.

\section{Hyper-parameter Study}
\label{sec:appendix_hyperparams}

\begin{figure*}[t]
    \centering
    \includegraphics[width=\linewidth]{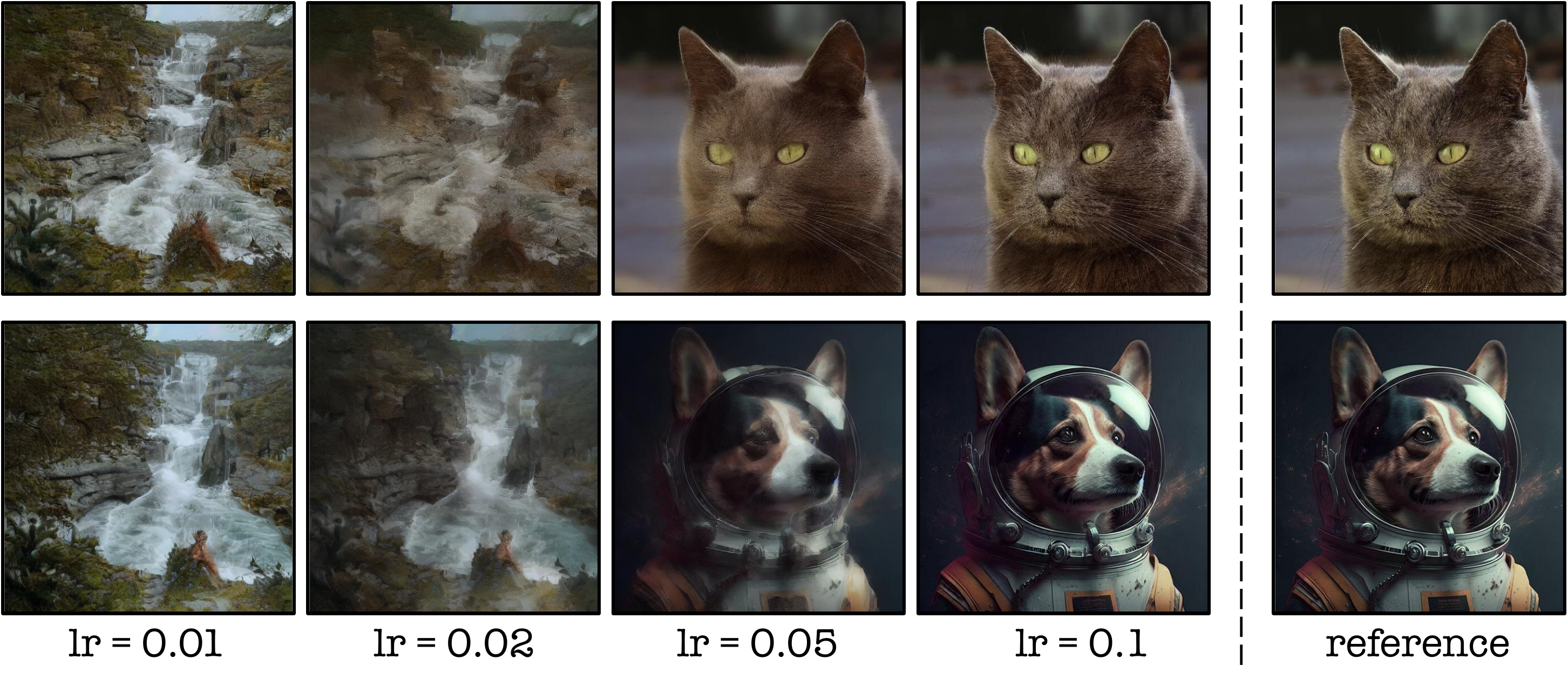}
    \caption{Effect of \flowchef~learning rate with fixed 20 max steps and one optimization step on InstaFlow.}
    \label{fig:hyperpara_lr}
\end{figure*}

\begin{figure*}[t]
    \centering
    \includegraphics[width=\linewidth]{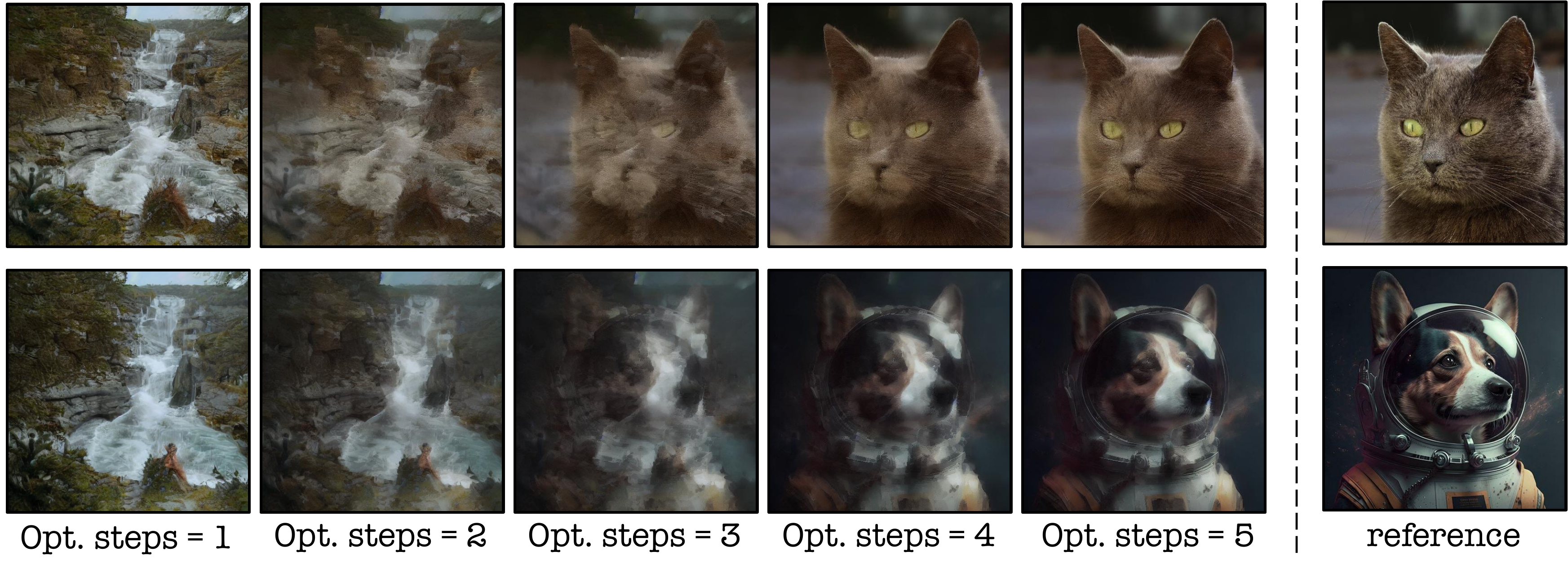}
    \caption{Effect of \flowchef~optimization steps with fixed 20 max steps and 0.02 learning rate on InstaFlow.}
    \label{fig:hyperpara_opt_steps}
\end{figure*}

\begin{figure*}[t]
    \centering
    \includegraphics[width=\linewidth]{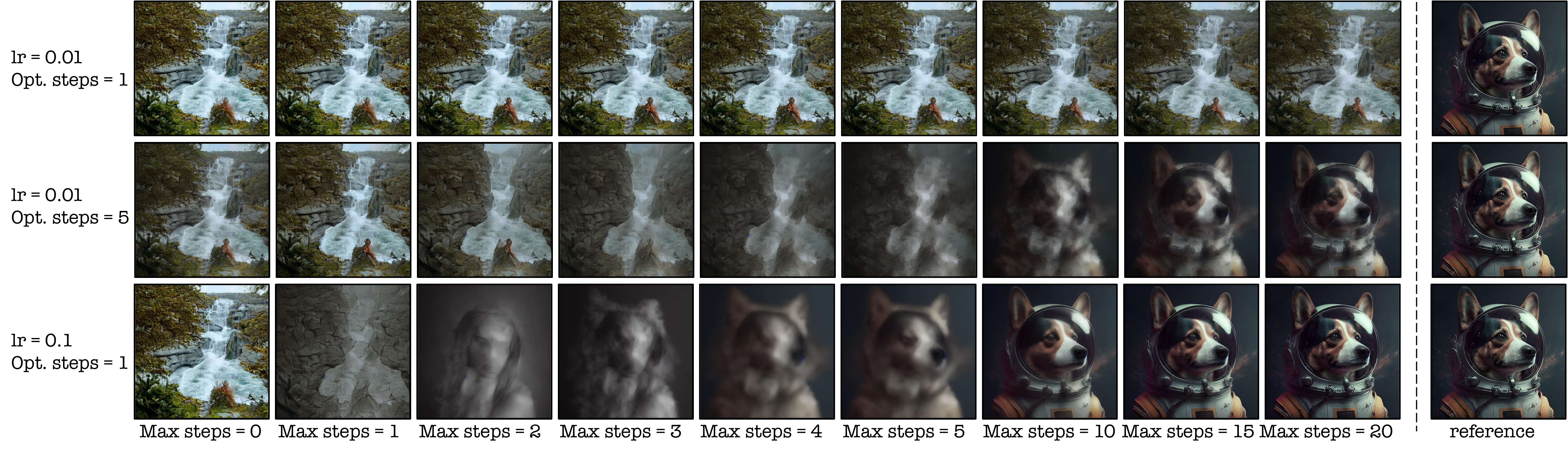}
    \caption{Effect of various \flowchef's steering parameters with increasing maximum optimization steps on InstaFlow.}
    \label{fig:hyperpara_all_max_steps}
\end{figure*}

Figures~\ref{fig:hyperpara_lr},~\ref{fig:hyperpara_opt_steps}, and~\ref{fig:hyperpara_all_max_steps} present an analysis of the impact of various hyperparameters on steering the InstaFlow model using \flowchef. 
Figure~\ref{fig:hyperpara_lr} demonstrates that a lower learning rate combined with a single optimization step is insufficient to effectively steer the model. Optimal performance is achieved with a learning rate of 0.1. Additionally, Figure~\ref{fig:hyperpara_opt_steps} shows that lower learning rates necessitate more optimization steps to achieve convergence.
Finally, Figure~\ref{fig:hyperpara_all_max_steps} illustrates how the denoising trajectory can be controlled by adjusting the learning rate and optimization steps, enabling recovery of the target sample with the desired accuracy. This control is particularly critical for image editing tasks, where striking the right balance between preserving the reference sample and applying the editing prompt is essential.
Table~\ref{tab:hyperpara_edit} further highlights optimal hyperparameter settings for image editing tasks, providing valuable guidance for achieving high-quality edits. This study underscores the flexibility of \flowchef~in adapting to diverse use cases by tuning these parameters effectively.

\begin{figure}[t]
    \centering
    \includegraphics[width=\linewidth]{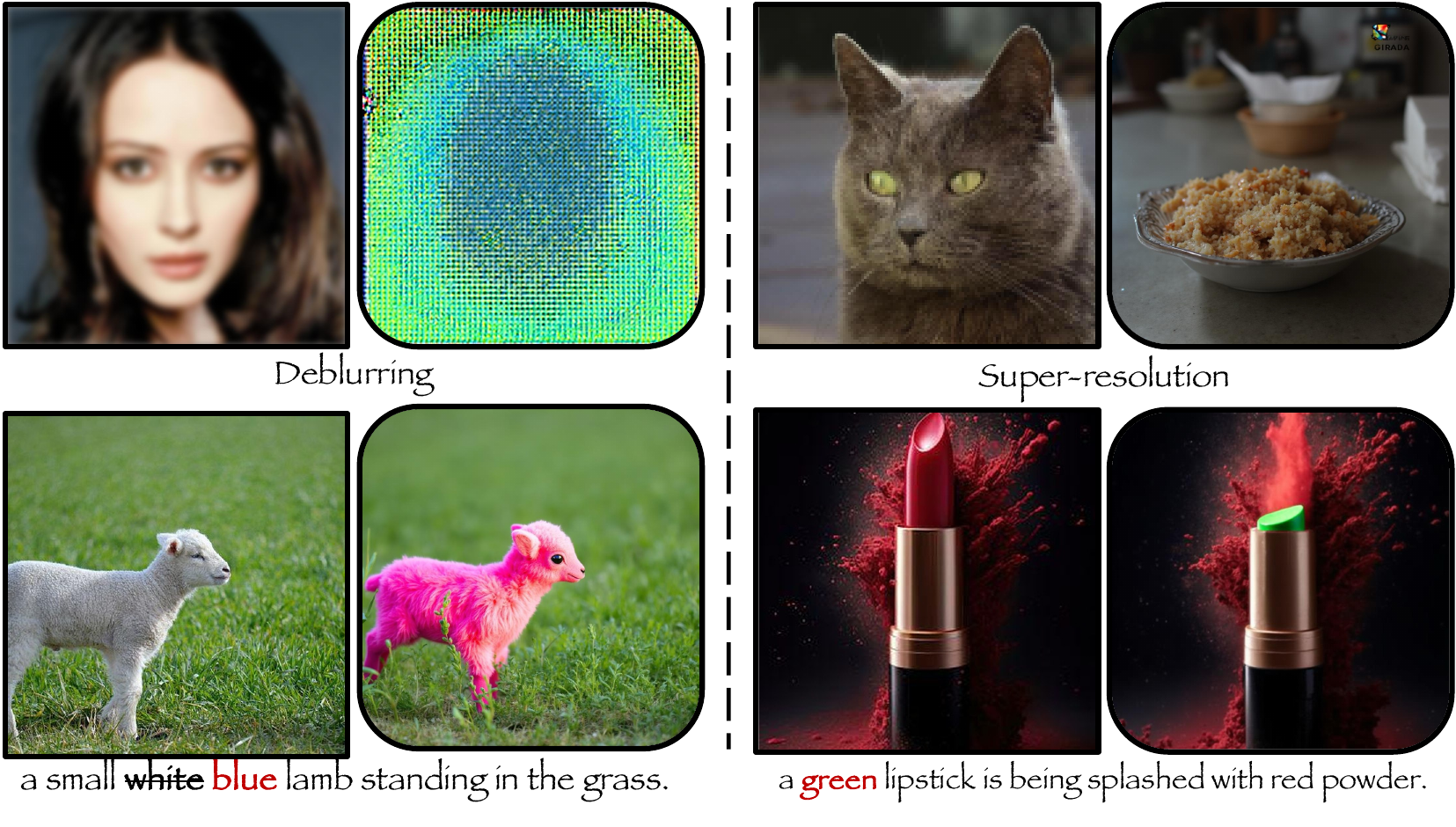}
    \caption{\flowchef~(Flux) model failures on inverse problems and image editing.}
    \label{fig:appendix_failure}
\end{figure}

\section{Qualitative Results}
\label{sec:appendix_qualitatives}

\begin{figure*}[t]
    \centering
    \includegraphics[width=\linewidth]{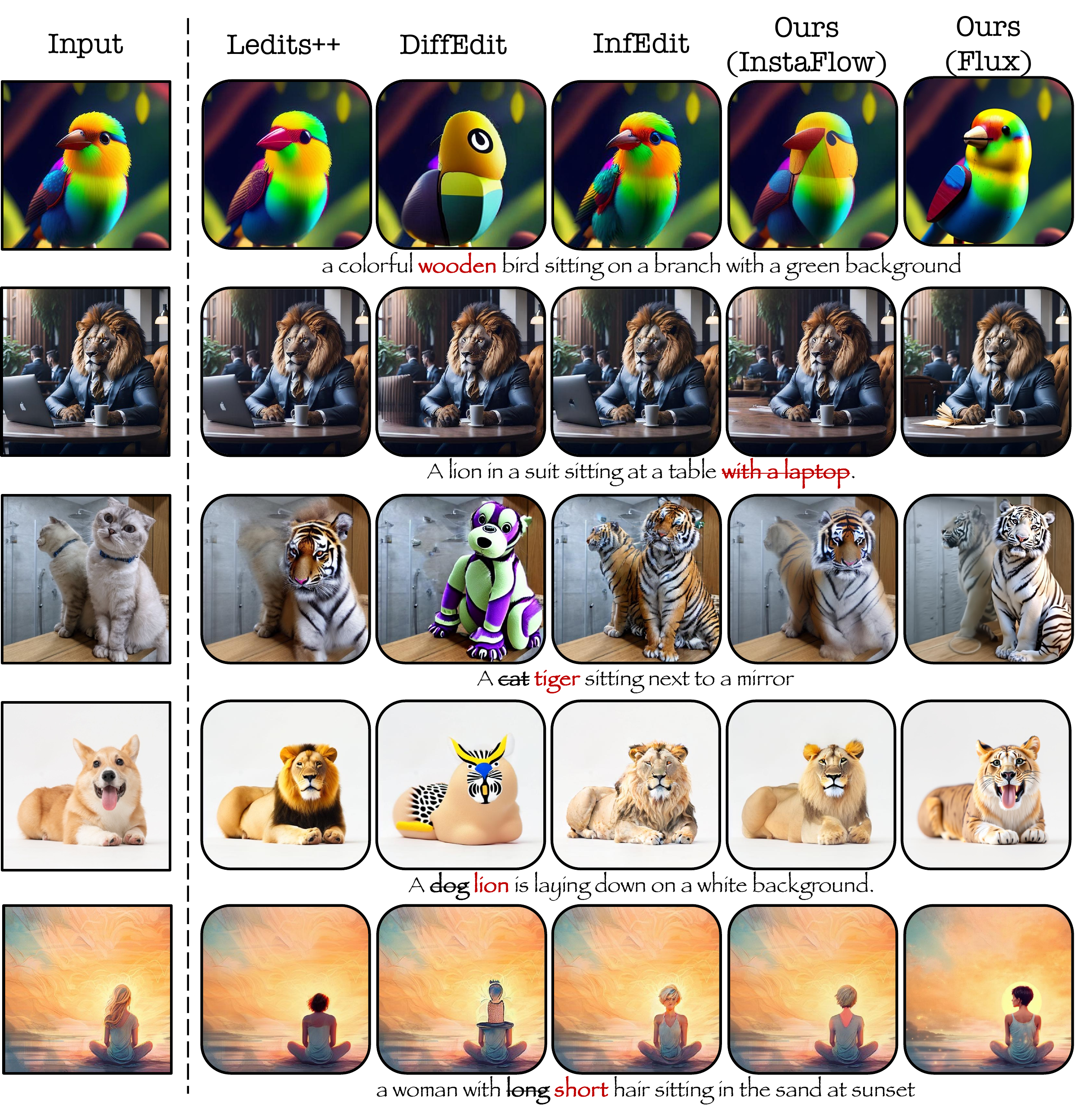}
    \caption{\textbf{Qualitative results on image editing.} Additional qualitative comparisons of \flowchef~with the baselines.}
    \label{fig:appendix_editing}
\end{figure*}

Figure~\ref{fig:appendix_editing} showcases additional qualitative examples of image editing tasks. For tasks such as changing materials or removing objects, \flowchef~outperforms the baselines significantly. However, some limitations are noted: while \flowchef~(InstaFlow) struggles to replace a cat with a tiger, InfEdit handles this task effectively, and Ledits++ exhibits difficulties. On the other hand, \flowchef~(Flux) achieves superior results, though it replaces a dog with a tiger instead of a lion in one instance. In the final example, both Ledits++ and \flowchef~successfully edit long hair into short hair. Importantly, the results in Figure~\ref{fig:appendix_editing} are presented without cherry-picking, using consistent hyperparameters for both baselines and \flowchef. Variability in outcomes may still arise due to random seeds and fine-tuned hyperparameter selection.

\begin{figure*}[t]
    \centering
    \includegraphics{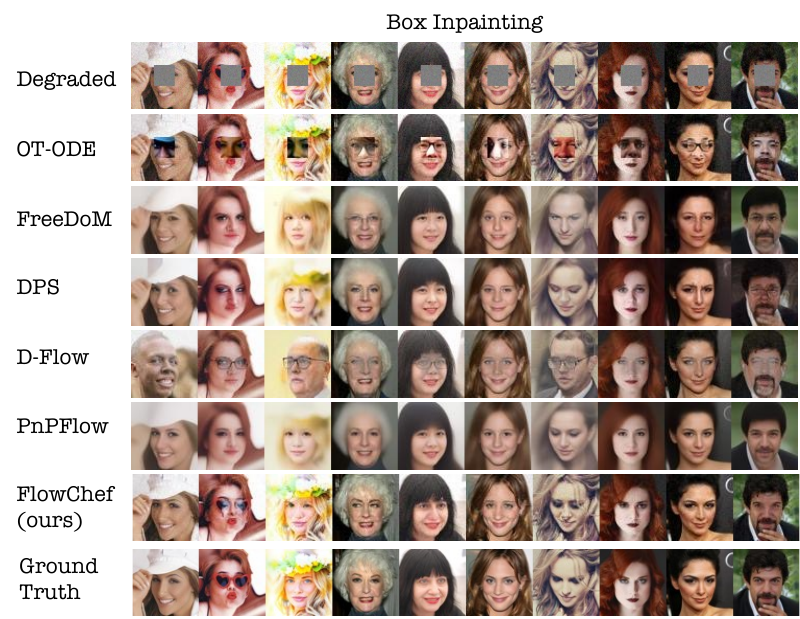}
    \caption{Qualitative examples of various methods for easy box inpainting task on RF++.}
    \label{fig:easy_inpaint}
\end{figure*}

\begin{figure*}[t]
    \centering
    \includegraphics{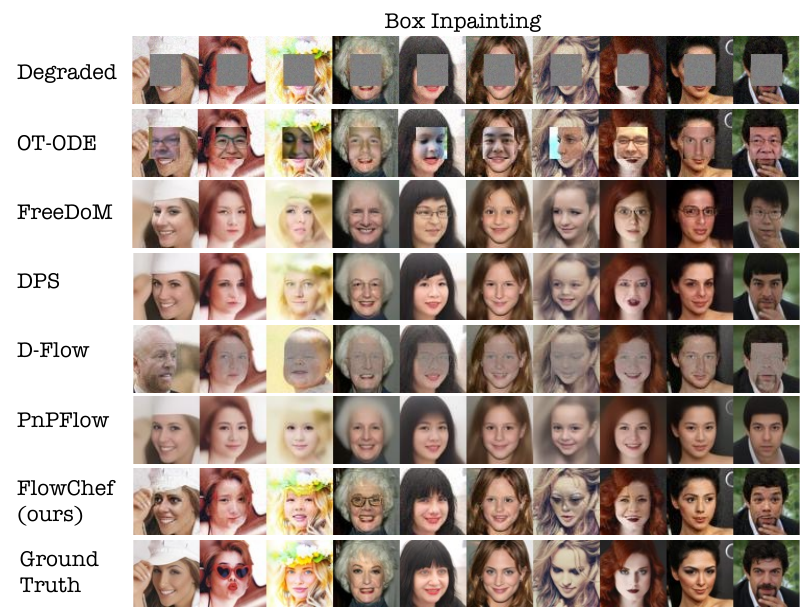}
    \caption{Qualitative examples of various methods for hard box inpainting task on RF++.}
    \label{fig:hard_inpaint}
\end{figure*}

\begin{figure*}[t]
    \centering
    \includegraphics{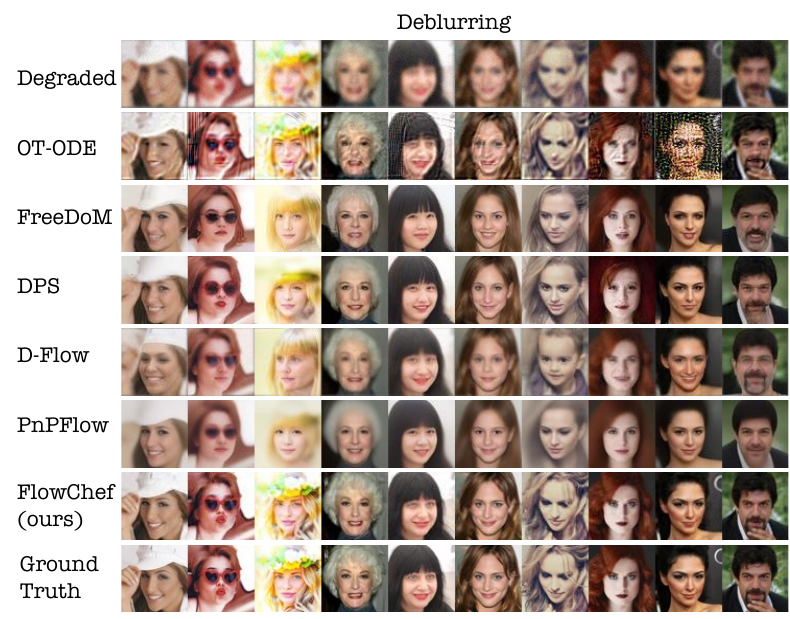}
    \caption{Qualitative examples of various methods for an easy deblurring task on RF++.}
    \label{fig:easy_deblur}
\end{figure*}

\begin{figure*}[t]
    \centering
    \includegraphics{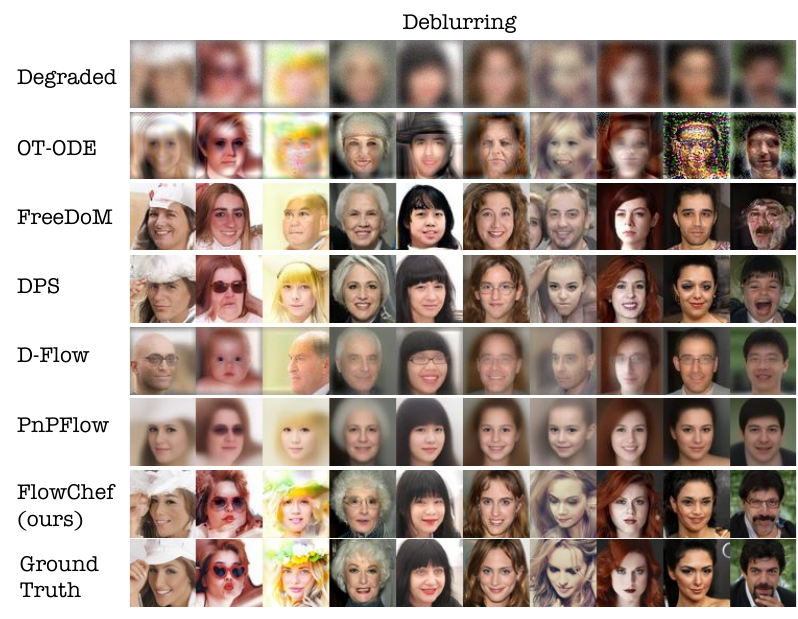}
    \caption{Qualitative examples of various methods for the hard deblurring task on RF++.}
    \label{fig:hard_deblur}
\end{figure*}

\begin{figure*}[t]
    \centering
    \includegraphics{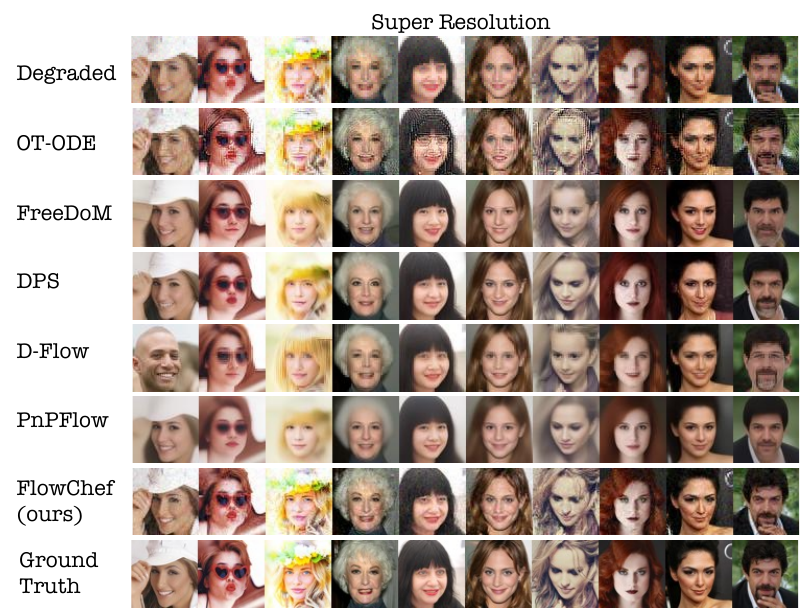}
    \caption{Qualitative examples of various methods for an easy super-resolution task on RF++.}
    \label{fig:easy_super}
\end{figure*}

\begin{figure*}[t]
    \centering
    \includegraphics{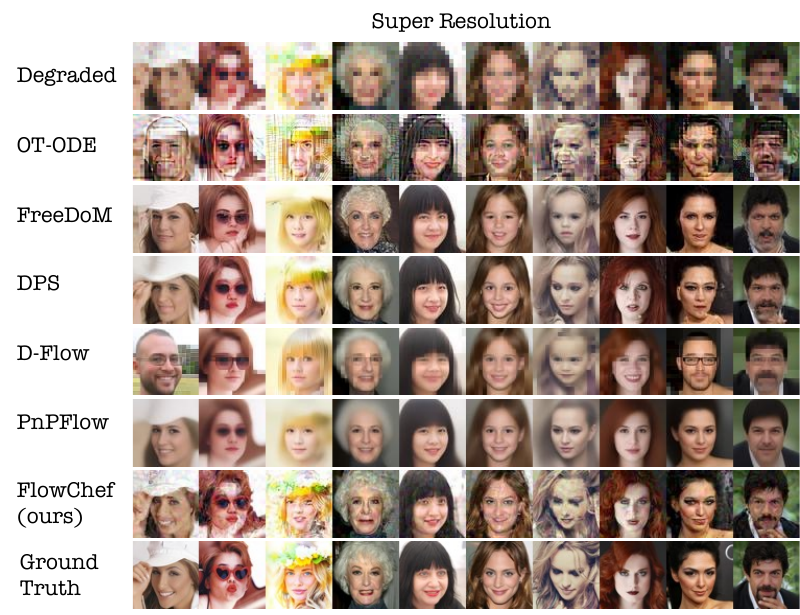}
    \caption{Qualitative examples of various methods for the hard super-resolution task on RF++.}
    \label{fig:hard_super}
\end{figure*}

Figures~\ref{fig:easy_inpaint},~\ref{fig:hard_inpaint},~\ref{fig:easy_deblur},~\ref{fig:hard_deblur},~\ref{fig:easy_super}, and~\ref{fig:hard_super} provide pixel-level qualitative results for various inverse problems, spanning inpainting, deblurring, and super-resolution tasks under both easy and hard scenarios. Readers are encouraged to zoom in to inspect these comparisons more closely. 
For each task, we randomly selected 10 CelebA examples and evaluated various baselines. Across all difficulty levels, FreeDoM, DPS, and PnPFlow demonstrate better performance than D-Flow and OT-ODE. However, \textbf{\flowchef~consistently outperforms all baselines}, producing sharp and visually appealing results where other methods either fail outright or introduce excessive smoothness. 
Hard scenarios pose challenges for all methods, but \flowchef~notably improves performance even under these conditions. While \flowchef~shows promise, future work is needed to address potential adversarial effects and further enhance robustness.

\section{Limitations \& Future Work}
\label{sec:appendix_futurework}

\paragraph{Limitations.}
While \flowchef~represents a significant leap in steering RFMs for controlled generation, it shares some limitations with its baseline counterparts. Hyperparameter tuning remains a challenge, particularly due to differences in trajectory behavior. For instance, while InstaFlow trajectories are relatively linear, Flux.1[Dev] trajectories exhibit non-linearity, necessitating careful tuning. 
As shown in Figure~\ref{fig:appendix_multiedit}, \flowchef~(Flux) faces difficulties in deblurring and super-resolution tasks, which we attribute to the pixel-space loss and non-linear behavior of the VAE model. Importantly, these limitations occur in less than 10\% of cases and can often be resolved by simply adjusting the random seed. Furthermore, due to Flux's lack of true classifier-free guidance (CFG), Algorithm~\ref{algo:algo3} occasionally fails to perfectly execute color changes, sometimes producing the unaltered target image without reflecting the edit (see Figure~\ref{fig:appendix_multiedit}).
Despite these minor limitations, \flowchef~still delivers state-of-the-art performance, making these challenges opportunities for further refinement rather than fundamental drawbacks.


\paragraph{Future Work.}
\flowchef~opens a promising avenue for steering RFMs effortlessly with guaranteed convergence for controlled image generation. While this work extensively evaluates \flowchef~on image generative models, future research should focus on expanding its capabilities to video and 3D generative models, areas that remain largely unexplored.
Additionally, the current implementation assumes the availability of human-annotated masks for image editing. Automating this step with advanced attention mechanisms could make \flowchef~a fully automated image editing framework. We encourage the research community to build upon this foundation to enhance its accessibility and functionality.

\paragraph{Ethical Concerns.}
As with all generative models, ethical concerns such as safety, misuse, and copyright issues apply to \flowchef~\cite{kim2024wouaf, kim2024race}. By enabling controlled generation with state-of-the-art RFMs, \flowchef~can be leveraged for beneficial and harmful purposes. To mitigate these risks, future efforts should focus on solutions such as image watermarking, content moderation, and unlearning harmful behaviors. While these issues are not unique to \flowchef, addressing them will be key to ensuring its responsible use.

\end{document}